\documentclass[10pt,twocolumn,letterpaper]{article}

\usepackage{cvpr}
\usepackage{times}
\usepackage{epsfig}
\usepackage{graphicx}
\usepackage{amsmath}
\usepackage{amssymb}
\usepackage{float}
\usepackage{multirow}
\usepackage{url}

\usepackage{makecell}
\usepackage{subcaption}
\usepackage{cuted}
\usepackage[font={small}]{caption}
\usepackage{pifont}
\usepackage[dvipsnames]{xcolor}
\usepackage{balance}

\usepackage{float}
\floatstyle{plaintop}
\restylefloat{table}

\definecolor{citecolor}{RGB}{65,105,225}
\usepackage[pagebackref=true,breaklinks=true,letterpaper=true,colorlinks,citecolor=citecolor,bookmarks=false]{hyperref}

\cvprfinalcopy 

\newcommand{\tabincell}[2]{\begin{tabular}{@{}#1@{}}#2\end{tabular}}

\def\cvprPaperID{****} 

\newcommand\blfootnote[1]{%
  \begingroup
  \renewcommand\thefootnote{}\footnote{#1}%
  \addtocounter{footnote}{-1}%
  \endgroup
}

\begin{document}

\title{DeeperForensics-1.0: A Large-Scale Dataset for \\Real-World Face Forgery Detection
\vspace{-0.16in}
}

\author{Liming Jiang$^{1}$ \hspace{2pt} Ren Li$^{2}$ \hspace{2pt} Wayne Wu$^{1,2}$ \hspace{2pt} Chen Qian$^{2}$ \hspace{2pt} Chen Change Loy$^{1\dagger}$\\
$^1$Nanyang Technological University \hspace{12pt} $^2$SenseTime Research\\
{\tt\small liming002@ntu.edu.sg} \hspace{0.8cm}
{\tt\small tomo.blade.lee@hotmail.com} \hspace{0.4cm}\\[-0.8pt]
{\tt\small wuwenyan@sensetime.com} \hspace{0.4cm}
{\tt\small qianchen@sensetime.com} \hspace{0.4cm}
{\tt\small ccloy@ntu.edu.sg}
}

\maketitle

\begin{strip}\centering
\vspace{-0.6in}
\includegraphics[width=\textwidth]{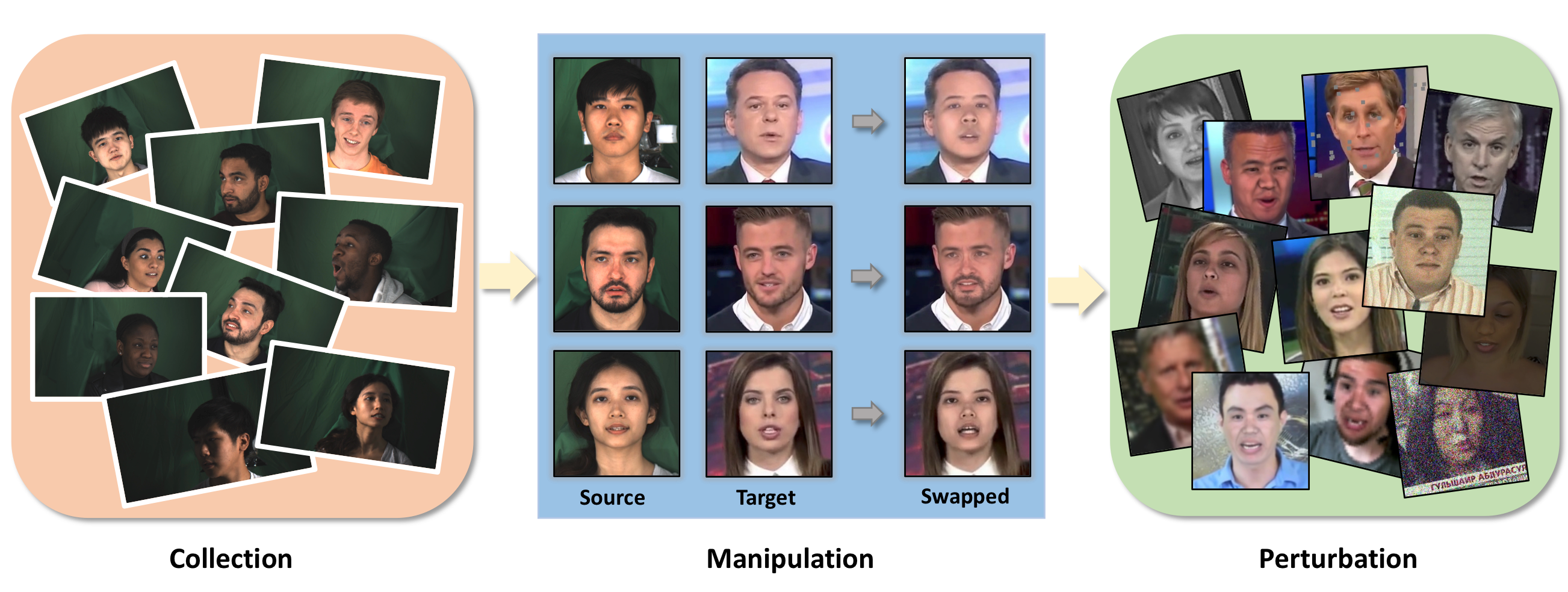}
 \vspace{-.26in}
\captionof{figure}{DeeperForensics-$1.0$ dataset is a new large-scale dataset for \textit{real-world} face forgery detection.}
\label{fig:first_figure}
\end{strip}

\begin{abstract}

We present our on-going effort of constructing a large-scale benchmark for face forgery detection.
The first version of this benchmark, DeeperForensics-$1.0$, represents the largest face forgery detection dataset by far, with $60,000$ videos constituted by a total of $17.6$ million frames, $10$ times larger than existing datasets of the same kind. Extensive real-world perturbations are applied to obtain a more challenging benchmark of larger scale and higher diversity.
All source videos in DeeperForensics-$1.0$ are carefully collected, and fake videos are generated by a newly proposed end-to-end face swapping framework. The quality of generated videos outperforms those in existing datasets, validated by user studies.
The benchmark features a hidden test set, which contains manipulated videos achieving high deceptive scores in human evaluations.
We further contribute a comprehensive study that evaluates five representative detection baselines and make a thorough analysis of different settings.
\footnote{\hspace{0.07cm}GitHub: \href{https://github.com/EndlessSora/DeeperForensics-1.0}{https://github.com/EndlessSora/DeeperForensics-1.0}.}$^,$\footnote{\hspace{0.07cm}Project page: \href{https://liming-jiang.com/projects/DrF1/DrF1.html}{https://liming-jiang.com/projects/DrF1/DrF1.html}.}\blfootnote{\hspace{-0.14cm}$^{\dagger}$ Corresponding author.}

\end{abstract}

\section{Introduction}

Face swapping has become an emerging topic in computer vision and graphics. Indeed, many works~\cite{DFL, DeepFakes, faceswap-GAN} on automatic face swapping have been proposed in recent years. These efforts have circumvented the cumbersome and tedious manual face editing processes, hence expediting the advancement in face editing. At the same time, such enabling technology has sparked legitimate concerns, particularly on its potential for being misused and abused. The popularization of ``Deepfakes'' on the internet has further set off alarm bells among the general public and authorities, in view of the conceivable perilous implications. Accordingly, there is a dire need for countermeasures to be in place promptly, particularly innovations that can effectively detect videos that have been manipulated.


Working towards forgery detection, various groups have contributed datasets (\eg, FaceForensics++~\cite{FF++data}, Deep Fake Detection~\cite{google} and DFDC~\cite{DFDC}) comprising manipulated video footages. The availability of these datasets has undoubtedly provided essential avenues for research into forgery detection. Nonetheless, the aforementioned datasets suffer several drawbacks. Videos in these datasets are either of a small number, of low quality, or overly artificial. Understandably, these datasets are inadequate to train a good model for effective forgery detection in \textit{real-world} scenarios. This is particularly true when current advances in human face editing are able to produce extremely realistic videos, rendering forgery detection a highly challenging task. On another note, we observe high similarity between training and test videos, in terms of their distribution, in certain works \cite{celebdf,FF++data}. Their actual efficacy in detecting \textit{real-world} face forgery cases, which are much more variable and unpredictable, remains to be further elucidated.

We believe that forgery detection models can only be enhanced when trained with a dataset that is exhaustive enough to encompass as many potential real-world variations as possible. To this end, we propose a large-scale dataset named DeeperForensics-$1.0$ consisting of $60,000$ videos with a total of $17.6$ million frames for real-world face forgery detection. The main steps of our dataset construction are shown in Figure~\ref{fig:first_figure}. We set forth three yardsticks when constructing this dataset: 1) \textit{Quality.} The dataset shall contain videos more realistic and much closer to the distribution of real-world detection scenarios. (Section~\ref{datacollection} and \ref{DFVAE}) 2) \textit{Scale.} The dataset shall be made up of a large-scale video sets. (Section~\ref{scaleanddiversity}) 3) \textit{Diversity.} There shall be sufficient variations in the video footages (\eg, compression, blurry, transmission errors) to match those that may be encountered in the real world (Section~\ref{scaleanddiversity}).

The primary challenge in the preparation of this dataset is the lack of good-quality video footages. Specifically, most publicly available videos are shot under an unconstrained environment resulting in large variations, including but not limited to suboptimal illumination, large occlusion of the target faces, and extreme head poses. Importantly, the lack of official informed consents from the video subjects precludes the use of these videos, even for non-commercial purposes. On the other hand, while some videos of manipulated faces are deceptively real, a larger number remains easily distinguishable by human eyes. The latter is often caused by model negligence towards appearance variations or temporal differences, leading to preposterous and incongruous results.

We approach the aforementioned challenge from two perspectives. 
1) Collecting fresh face data from $100$ individuals with informed consents (Section~\ref{datacollection}).
2) Devising a novel method, DeepFake Variational Auto-Encoder (DF-VAE), to enhance existing videos (Section~\ref{DFVAE}).
In addition, we introduce diversity into the video footages through deliberate addition of distortions and perturbations, simulating real-world scenarios. We collate the newly collected data and the DF-VAE-modified videos into the DeeperForensics-$1.0$ dataset, with the aim of further expanding it gradually over time. We benchmark five representative open-source forgery detection methods using our dataset as well as a hidden test set containing manipulated videos that achieve high deceptive ranking in user studies.

We summarize our contributions as follows: 1) We propose a new dataset, DeeperForensics-$1.0$ that is larger in scale than existing ones, of high quality and rich diversity. To improve its quality, we introduce a carefully designed data collection and a novel framework, DF-VAE, that effectively mitigate obvious fabricated effects of existing manipulated videos. DeeperForensics-$1.0$ dataset shall facilitate future research in forgery detection of human faces in real-world scenarios. 2) We benchmark results of existing representative forgery detection methods on our dataset, offering insights into the current status and future strategy in face forgery detection.

\begin{table*}
\small
\centering
\begin{tabular}{c|c|c|c|c|c|c|c}
\Xhline{1.2pt}
Dataset  & Total videos & \tabincell{c}{Ratio\\(real : fake)}  &  
\tabincell{c}{Controlled\\Capture} & 
\tabincell{c}{Consented\\Actors} &
\tabincell{c}{Perturbations\\(total number)} & 
\tabincell{c}{Perturbations\\(mixture)} &
\tabincell{c}{New\\Method} \\
\Xhline{1.2pt}
UADFV \cite{UADFV}  & 98  &  1 : 1  & $\times$  & -- & -- & $\times$ & $\times$ \\
DeepFake-TIMIT \cite{DFTIMIT} &620& only fake & $\times$ &-- & --& $\times$ & $\times$ \\
Celeb-DF \cite{celebdf} & 1203  &  1 : 1.95 & $\times$ & --& -- & $\times$ & $\times$ \\
FaceForensics++ \cite{FF++data} & 5000 & 1 : 4 & $\times$ & -- & 2 & $\times$ & $\times$ \\
\tabincell{c}{Deep Fake Detection \cite{google} \\
(joins FaceForensics++)}
       & 3431  &  1 : 8.5 & $\times$ & 28  & -- & $\times$ & $\times$ \\
DFDC Preview Dataset \cite{DFDC} & 5214 &  1 : 3.6 & $\times$ & 66 &  3 & $\times$ & $\times$ \\
\Xhline{1.2pt}
\textbf{DeeperForensics-1.0 (Ours)}  & \textbf{60000} &  5 : 1 & \ding{52} & \textbf{100} &  \textbf{35} & \ding{52} & \ding{52} \\
\Xhline{1.2pt}
\end{tabular}
\caption{The most relevant datasets compared to our dataset. DeeperForensics-1.0 is an order of magnitude larger in scale than existing datasets \textit{w.r.t.} both real and fake parts. We build a professional indoor environment to better control the important attributes of the collected data. $100$ paid actors give consents to the use and manipulation of their faces by signing a formal agreement. We employ seven types of perturbations at five intensity levels, leading to $35$ perturbations in total. The video may be subjected to a mixture of more than one perturbation. In contrast to prior works, we also introduce a new end-to-end high-fidelity face swapping method.}
 \vspace{-.05in}
\label{tab:datasetdiff}
\end{table*}

\section{Related Work}
This paper includes two main aspects of face forgery detection related to other works: dataset and benchmark. We will cover some important works in this section.

\noindent
\textbf{Face forgery detection datasets.} Building a dataset for forgery detection requires a huge amount of effort on data collection and manipulation. Early forgery detection datasets comprise images captured under highly restrictive conditions, \eg, MICC\_F2000 \cite{miccf2000}, Wild Web dataset \cite{wildweb}, Realistic Tampering dataset \cite{rtddataset}. 


Owing to the urgency in video-based face forgery detection, some prominent groups have devoted their efforts to create face forensics video datasets (see Table \ref{tab:datasetdiff}). UADFV \cite{UADFV} contains $98$ videos, \ie, $49$ real videos from YouTube and $49$ fake ones generated by FakeAPP \cite{fakeapp}. DeepFake-TIMIT \cite{DFTIMIT} manually selects $16$ similar looking pairs of people from VidTIMIT \cite{VIDTIMIT} database. For each of the $32$ subjects, they generate about $10$ videos using low-quality and high-quality versions of faceswap-GAN \cite{faceswap-GAN}, resulting in a total of $620$ fake videos. Celeb-DF \cite{celebdf} includes $408$ YouTube videos, mostly of celebrities, from which $795$ fake videos are synthesized. FaceForensics++ \cite{FF++data} is the first large-scale face forensic dataset that consists of $4,000$ fake videos manipulated by four methods (\ie, DeepFakes \cite{DeepFakes}, Face2Face \cite{face2face}, FaceSwap \cite{FaceSwap}, NeuralTextures \cite{NeuralTextures})), and $1,000$ real videos from YouTube. Afterwards, Google joins FaceForensics++ and contributes Deep Fake Detection \cite{google} dataset with $3,431$ real and fake videos from $28$ actors. Recently, Facebook invites $66$ individuals and builds the DFDC preview dataset \cite{DFDC}, which includes $5,214$ original and tampered videos with three types of augmentations.

In comparison, we invite $100$ paid actors and collect high-resolution ($1920 \times 1080$) source data with various poses, expressions, and illuminations. 3DMM blendshapes \cite{3dmm} are taken as reference to supplement some extremely exaggerated expressions. We get consents from all the actors for using and manipulating their faces. In contrast to prior works, we also propose a new end-to-end face swapping method (\ie, DF-VAE) and systematically apply seven types of perturbations to the fake videos at five intensity levels. The mixture of distortions to a single video makes our dataset better imitate real-world scenarios. Ultimately, we construct DeeperForensics-$1.0$ dataset, which contains up to $60,000$ high-quality videos with a total of $17.6$ million frames.

\noindent
\textbf{Face forgery detection benchmarks.} 
%
A new prominent benchmark, FaceForensics Benchmark \cite{FF++data}, for facial manipulation detection has been proposed recently. The benchmark includes six image-level face forgery detection baselines \cite{mosonet, FF++bm3, xception, FF++bm2, FF++bm1, FF++bm4}. Although FaceForensics Benchmark adds distortions to the videos by converting them into different compression rates, a deeper exploration of more perturbation types and their mixture is missing. Celeb-DF \cite{celebdf} also provides a face forgery detection benchmark including seven methods \cite{mosonet, xception, cdfbm2, cdfbm3, cdfbm4, UADFV, cdfbm1} trained and tested on different datasets.
In aforementioned benchmarks, the test set usually shares a similar distribution with the training set. Such an assumption inherently introduces biases and renders these methods impractical for face forgery detection in real-world settings with much more diverse and unknown fake videos.

In our benchmark, we introduce a challenging hidden test set with manipulated videos that achieve high deceptive scores in user studies, to better simulate \textit{real-world} distribution. Various perturbations are analyzed to make our benchmark more comprehensive. In addition, we mainly exploit \textit{video-level} forgery detection baselines \cite{i3d,resnet,lstm,c3d,tsn}. Temporal information -- a significant cue for video forgery detection besides single-frame quality -- has been considered. We will elaborate our benchmark in Section~\ref{benchmark}.

\section{A New Large-Scale Face Forensics Dataset}
\label{dataset}
The main contribution of this paper is a new large-scale dataset for real-world face forgery detection, DeeperForensics-$1.0$, which provides an alternative to existing databases. DeeperForensics-$1.0$ consists of $60,000$ videos with $17.6$ million frames in total, including $50,000$ original collected videos and $10,000$ manipulated videos. To construct a dataset more suitable for real-world face forgery detection, we design this dataset with careful consideration of \textit{quality}, \textit{scale}, and \textit{diversity}. In Section~\ref{datacollection} and~\ref{DFVAE}, we will discuss the details of data collection and methodology (\ie, DF-VAE) to improve \textit{quality}. In Section~\ref{scaleanddiversity}, we will show how to ensure large \textit{scale} and high \textit{diversity} of DeeperForensics-$1.0$.


\subsection{Data Collection}
\label{datacollection}

\begin{figure}
   \begin{center}
       \includegraphics[width=\linewidth]{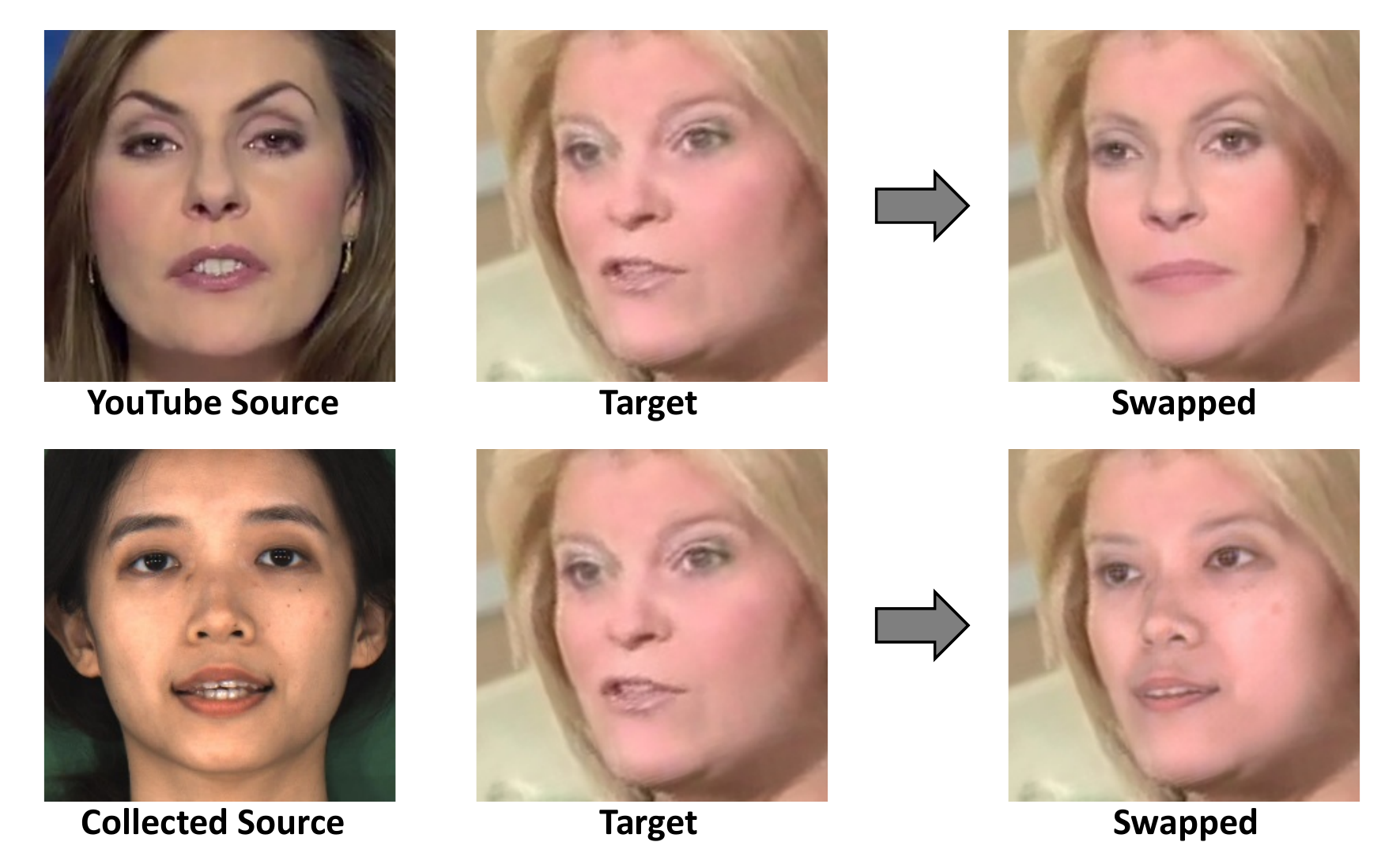}
   \end{center}
   \vspace{-0.45cm}
      \caption{Comparison of using only YouTube video and the collected video as source data, with the same method and setting.}
   \vspace{-0.35cm}
   \label{fig:data_collection_compare}
\end{figure}

\begin{figure*}
   \begin{center}
       \includegraphics[width=\linewidth]{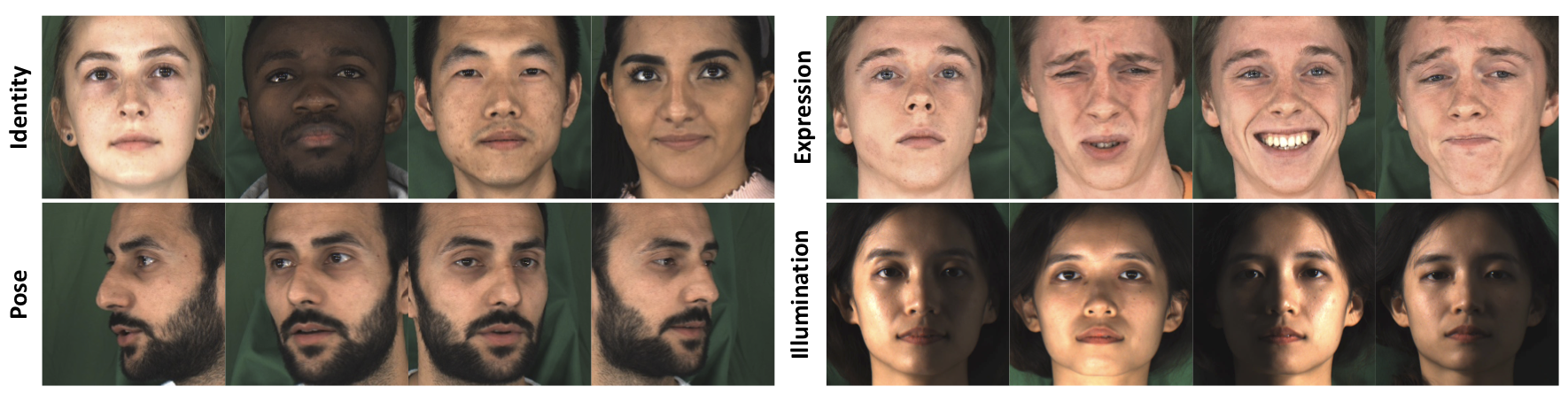}
   \end{center}
   \vspace{-0.45cm}
      \caption{Diversity in identities, poses, expressions, and illuminations in our collected source data.}
   \vspace{-0.15cm}
   \label{fig:data_collection}
\end{figure*}

Source data is the first factor that highly affects \textit{quality}. Taking results in Figure~\ref{fig:data_collection_compare} as an example, the source data collection increases the robustness of our face swapping method to extreme poses, since videos on the internet usually have limited head pose variations.

We refer to the identity in the driving video as the ``target'' face and the identity of the face that is swapped onto the driving video as the ``source'' face. Different from previous works, we find that the source faces play a much more critical role than the target faces in building a high-quality dataset. Specifically, the expressions, poses, and lighting conditions of source faces should be much richer in order to perform robust face swapping. Hence, our data collection mainly focuses on source face videos. Figure~\ref{fig:data_collection} shows the diversity in different attributes of our data collection.

We invite $100$ paid actors to record the source videos. Similar to \cite{google, DFDC}, we obtain consents from all the actors for using and manipulating their faces to avoid the portrait right issues. The participants are carefully selected to ensure variability in genders, ages, skin colors, and nationalities. We maintain a roughly equal proportion \textit{w.r.t.} each of the attributes above. In particular, we invite $55$ males and $45$ females from $26$ countries. Their ages range from $20$ to $45$ years old to match the most common age group appearing on real-world videos. The actors have four typical skin tones: \textit{white}, \textit{black}, \textit{yellow}, \textit{brown}, with ratio $1$:$1$:$1$:$1$. All faces are clean without glasses or decorations.

\begin{figure}
   \begin{center}
       \includegraphics[width=\linewidth]{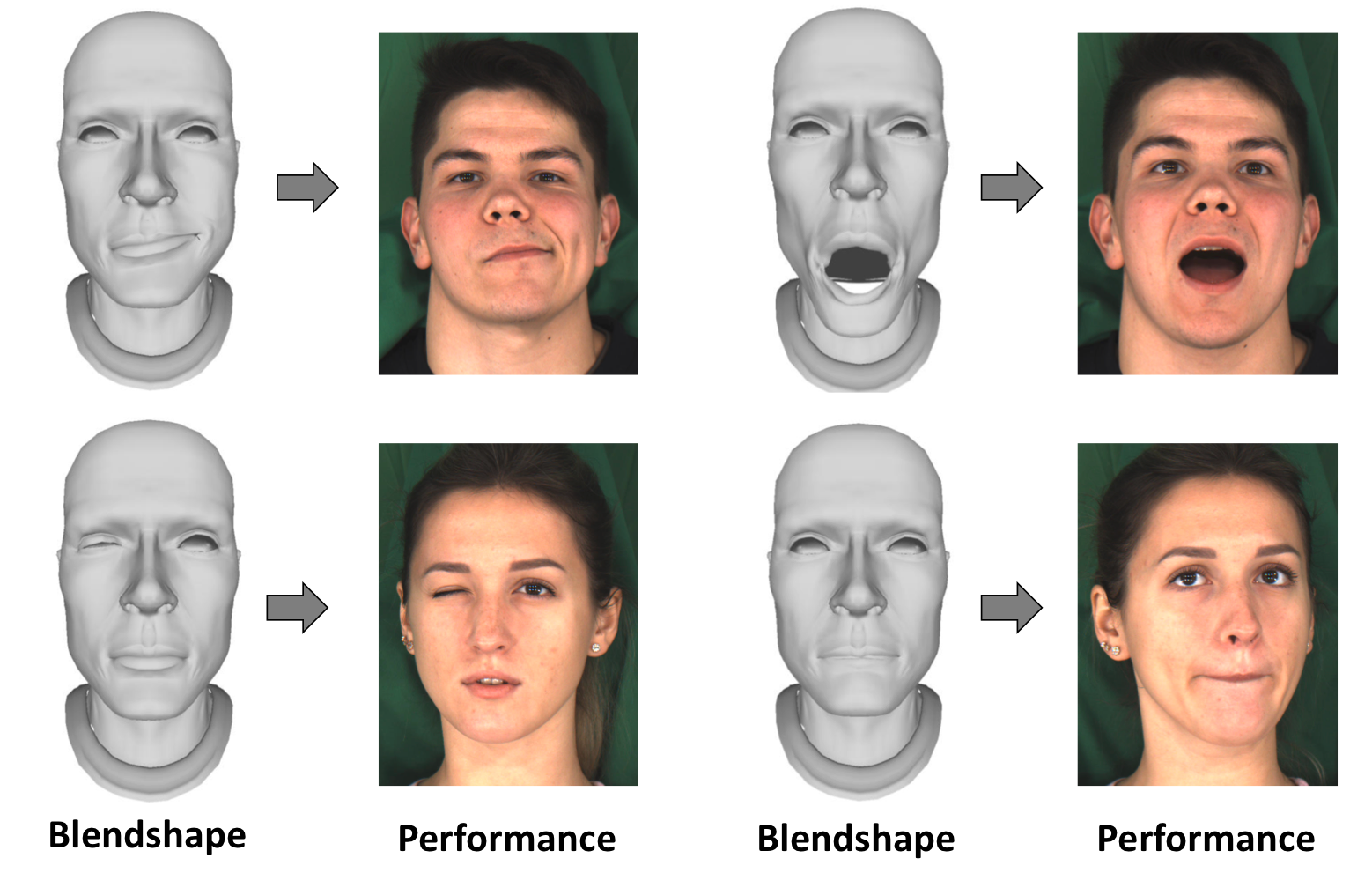}
   \end{center}
   \vspace{-0.35cm}
      \caption{Examples of 3DMM blendshapes in our data collection.}
   \vspace{-0.45cm}
   \label{fig:blendshape}
\end{figure}

Different from previous data collection in the wild (see Table~\ref{tab:datasetdiff}), we build a professional \textit{indoor} environment for a more controllable data collection. We only use the facial regions (detected and cropped by LAB \cite{wayne2018lab}) of the source data, so we can neglect the background. We set seven HD cameras from different angles: front, left, left-front, right, right-front, oblique-above, oblique-below. The resolution of our recorded videos is high ($1920 \times 1080$). We train the actors in advance to keep the collection process smooth. We request the actors to turn their heads and speak naturally with eight expressions: neutral, angry, happy, sad, surprise, contempt, disgust, fear. The head poses range from $-90^{\circ}$ to $+90^{\circ}$. Furthermore, the actors are asked to perform $53$ expressions defined in 3DMM blendshapes~\cite{3dmm} (see Figure~\ref{fig:blendshape}) to supplement some extremely exaggerated expressions. When performing 3DMM blendshapes, the actors also speak naturally to avoid excessive frames that show a closed mouth.

In addition to expressions and poses, we systematically set nine lighting conditions from various directions: uniform, left, top-left, bottom-left, right, top-right, bottom-right, top, bottom. The actors are only asked to turn their heads under uniform illumination, so the lighting remains unchanged on specific facial regions to avoid many duplicated data samples recorded by the cameras set at different angles.
In the end, our collected data contain over $50,000$ videos with a total of $12.6$ million frames -- an order of magnitude more than existing datasets. 

\begin{figure}[t]
   \begin{center}
       \includegraphics[width=\linewidth]{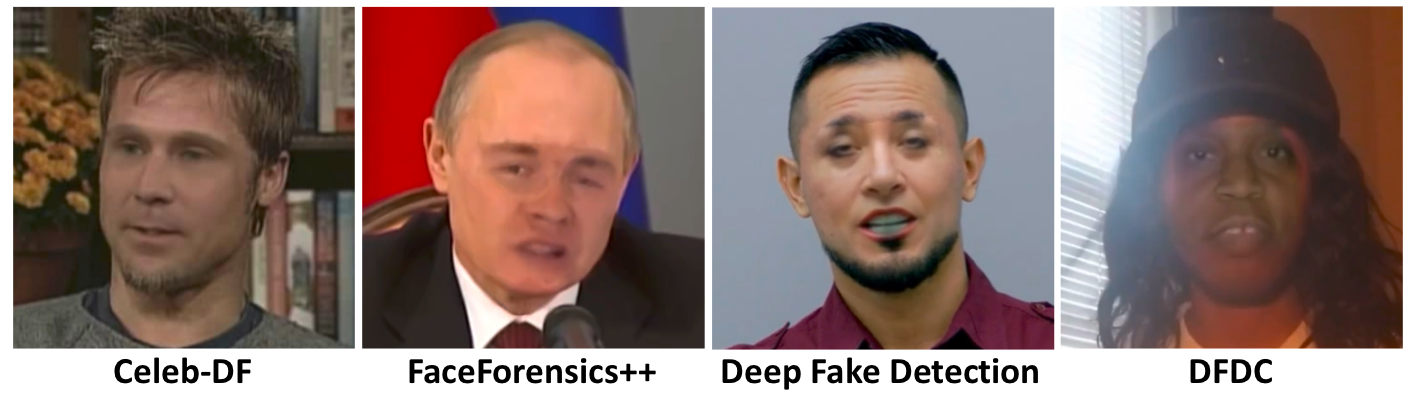}
   \end{center}
   \vspace{-0.35cm}
      \caption{Examples of style mismatch problems in prominent face forensics datasets.}
   \vspace{-0.35cm}
   \label{fig:stype_mismatch}
\end{figure}

\begin{figure*}[t!]
   \begin{center}
       \includegraphics[width=\linewidth]{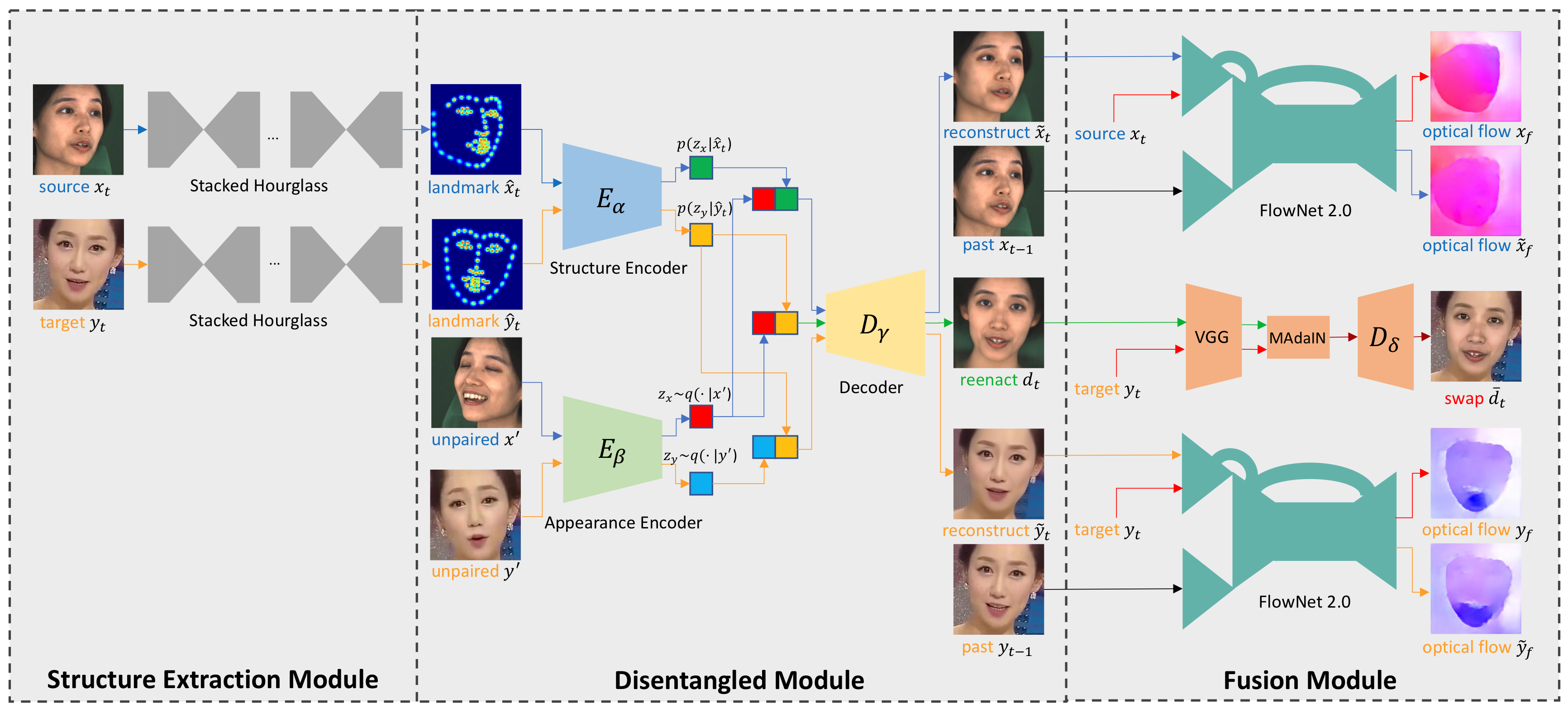}
   \end{center}
   \vspace{-0.35cm}
      \caption{The main framework of DeepFake Variational Auto-Encoder. In training, we reconstruct the source and target faces in blue and orange arrows, respectively, by extracting landmarks and constructing an unpaired sample as the condition. Optical flow differences are minimized after reconstruction to improve temporal continuity. In inference, we swap the latent codes and get the reenacted face in green arrows. Subsequent MAdaIN module fuses the reenacted face and the original background resulting in the swapped face.}
   \label{fig:main_framework}
   \vspace{-0.25cm}
\end{figure*}

\subsection{DeepFake Variational Auto-Encoder}
\label{DFVAE}

To tackle low visual \textit{quality} problems of previous works, we consider three key requirements in formulating a high-fidelity face swapping method: 
1) It should be general and scalable for us to generate large number of videos with high quality.
2) The problem of face style mismatch caused by appearance variations need to be addressed. Some failure cases of existing methods are shown in Figure \ref{fig:stype_mismatch}.
3) Temporal continuity of generated videos should be taken into consideration. 

Based on the aforementioned requirements, we propose DeepFake Variational Auto-Encoder (DF-VAE), a novel learning-based face swapping framework. DF-VAE consists of three main parts, namely a structure extraction module, a disentangled module, and a fusion module.
We will give a brief and intuitive understanding of the DF-VAE framework below. Please refer to the \textit{Appendix} for detailed derivations and results.

\begin{figure*}[t]
   \begin{center}
       \includegraphics[width=\linewidth]{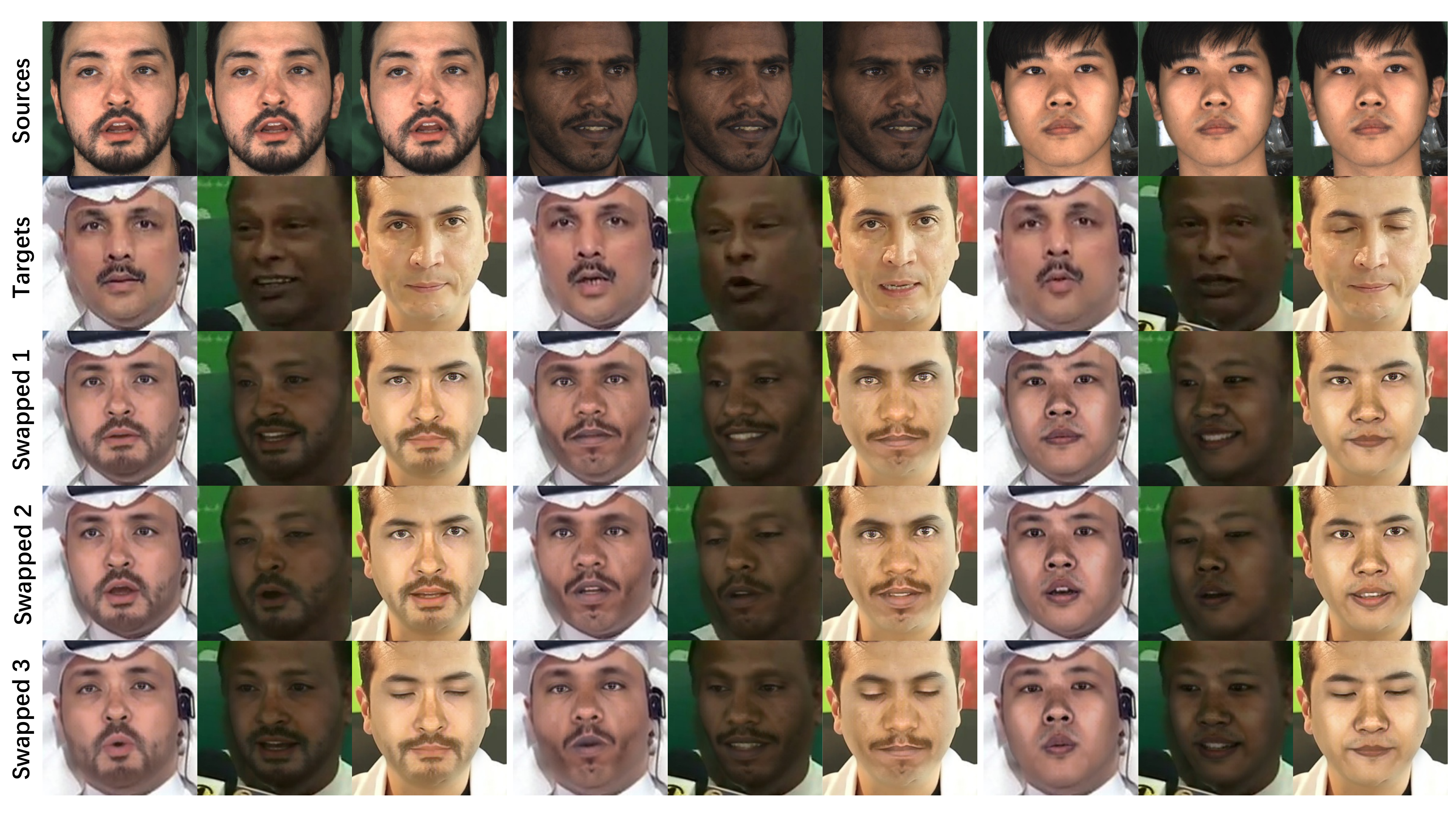}
   \end{center}
   \vspace{-0.35cm}
      \caption{Many-to-many (three-to-three) face swapping by a \textbf{single} model with obvious reduction of style mismatch problems. This figure shows the results between three source identities and three target identities. The whole process is end-to-end.}
   \label{fig:m2m}
   \vspace{-0.25cm}
\end{figure*}

\noindent
\textbf{Disentanglement of structure and appearance.}
The first step of our method is face reenactment -- animating the source face with similar expression as the target face, without any paired data. Face swapping is considered as a subsequent step of face reenactment that performs fusion between the reenacted face and the target background. For robust and scalable face reenactment, we should cleanly disentangle structure (\ie, expression and pose) and appearance representation (\ie, texture, skin color, \etc.) of a face. This disentanglement is rather difficult because structure and appearance representation are far from independent. We describe our solution as follows.

Let $\mathbf{x}_{1:T}\equiv{\{x_1,x_2,...,x_T\}}\in{X}$ be a sequence of source face video frames, and $\mathbf{y}_{1:T}\equiv{\{y_1,y_2,...,y_T\}}\in{Y}$ be the sequence of corresponding target face video frames. We first simplify our problem and only consider two specific snapshots at time $t$, $x_t$ and $y_t$. Let $\tilde{x}_t$, $\tilde{y}_t$, $d_t$ represent the reconstructed source face, the reconstructed target face, and the reenacted face, respectively.

Consider the reconstruction procedure of the source face $x_t$. Let $s_x$ denotes the structure representation and $a_x$ denotes the appearance information. The face generator can be depicted as the posteriori estimate $p_\theta\left(x_t|s_x,a_x\right)$. The solution of our reconstruction goal, marginal log-likelihood $\tilde{x}_t\sim\log{p_\theta\left(x_t\right)}$, by a common Variational Auto-Encoder (VAE) \cite{vae} can be written as:
\begin{equation}
\label{eq:1}
\begin{split}
    \log{p_\theta\left(x_t\right)}=D_{KL}\left(q_\phi\left(s_x,a_x|x_t\right)\|p_\theta\left(s_x,a_x|x_t\right)\right)\\
    +L\left(\theta,\phi;x_t\right),
\end{split}
\end{equation}
where $q_\phi$ is an approximate posterior to achieve the evidence lower bound (ELBO) in the intractable case, and the second RHS term $L\left(\theta,\phi;x_t\right)$ is the variational lower bound \textit{w.r.t.} both the variational parameters $\phi$ and generative parameters $\theta$. 

In Eq.~\eqref{eq:1}, we assume that both $s_x$ and $a_x$ are latent priors computed by the same posterior $x_t$. However, the separation of these two variables in the latent space is rather difficult without additional conditions. Therefore, we employ a simple yet effective approach to disentangle these two variables.

The blue arrows in Figure~\ref{fig:main_framework} demonstrate the reconstruction procedure of the source face $x_t$. Instead of feeding a single source face $x_t$, we sample another source face $x^\prime$ to construct unpaired data in the source domain. To make the structure representation more evident, we use the stacked hourglass networks \cite{stackedhourglass} to extract landmarks of $x_t$ in the structure extraction module and get the heatmap $\hat{x}_t$. Then we feed the heatmap $\hat{x}_t$ to the Structure Encoder $E_\alpha$, and $x^\prime$ to the Appearance Encoder $E_\beta$. We concatenate the latent representations (small cubes in red and green) and feed it to the Decoder $D_\gamma$. Finally, we get the reconstructed face $\tilde{x}_t$, \ie, marginal log-likelihood of $x_t$. 

Therefore, the latent structure representation $s_x$ in Eq.~\eqref{eq:1} becomes a more evident heatmap representation $\hat{x}_t$, which is introduced as a new condition. The unpaired sample $x^\prime$ with the same identity \textit{w.r.t.} $x_t$ is another condition, being a substitute for $a_x$. Eq.~\eqref{eq:1} can be rewritten as a conditional log-likelihood:
\begin{equation}
\label{eq:4}
\begin{split}
    \log{p_\theta\left(x_t|\hat{x}_t,x^\prime\right)}=D_{KL}\left(q_\phi\left(z_x|x_t,\hat{x}_t,x^\prime\right)\|p_\theta\left(z_x|x_t,\hat{x}_t,x^\prime\right)\right)\\
    +L\left(\theta,\phi;x_t,\hat{x}_t,x^\prime\right),
\end{split}
\end{equation}
The first RHS term KL-divergence is non-negative, we get:
\begin{equation}
\footnotesize
\label{eq:5}
\begin{split}
    &\log{p_\theta\left(x_t|\hat{x}_t,x^\prime\right)}\geq{L(\theta,\phi;x_t,\hat{x}_t,x^\prime)}\\
    &=\mathbb{E}_{q_\phi\left(z_x|x_t,\hat{x}_t,x^\prime\right)}\left[-\log{q_\phi\left(z_x|x_t,\hat{x}_t,x^\prime\right)}+\log{p_\theta\left(x_t,z_x|\hat{x}_t,x^\prime\right)}\right],
\end{split}
\end{equation}
and $L(\theta,\phi;x_t,\hat{x}_t,x^\prime)$ can also be written as:
\begin{equation}
\label{eq:6}
\begin{split}
    L\left(\theta,\phi;x_t,\hat{x}_t,x^\prime\right)=&-D_{KL}\left(q_\phi\left(z_x|x_t,\hat{x}_t,x^\prime\right)\|p_\theta\left(z_x|\hat{x}_t,x^\prime\right)\right)\\
    &+\mathbb{E}_{q_\phi\left(z_x|x_t,\hat{x}_t,x^\prime\right)}\left[\log{p_\theta\left(x_t|z_x,\hat{x}_t,x^\prime\right)}\right].
\end{split}
\end{equation}

We let the variational approximate posterior be a multivariate Gaussian with a diagonal covariance structure:
\begin{equation}
\label{eq:7}
\begin{split}
    \log{q_\phi\left(z_x|x_t,\hat{x}_t,x^\prime\right)}\equiv{\log{\mathcal{N}\left(z_x;\mathbf{\mu},\mathbf{\sigma^2}\mathbf{I}\right)}},
\end{split}
\end{equation}
where $\mathbf{I}$ is an identity matrix. Exploiting the reparameterization trick \cite{vae}, the non-differentiable operation of sampling can become differentiable by an auxiliary variable with independent marginal. In this case, $z_x\sim{q_\phi\left(z_x|x_t,\hat{x}_t,x^\prime\right)}$ is implemented by $z_x=\mu+\sigma\epsilon$ where $\epsilon$ is an auxiliary noise variable $\epsilon\sim\mathcal{N}(0,1)$. Finally, the approximate posterior $q_\phi(z_x|x_t,\hat{x}_t,x^\prime)$ is estimated by the separated encoders, Structure Encoder $E_\alpha$ and Appearance Encoder $E_\beta$, in an end-to-end training process by standard gradient descent.

We discuss the whole workflow of reconstructing the source face. In the target face domain, the reconstruction procedure is the same, as shown by orange arrows in Figure~\ref{fig:main_framework}. 

During training, the network learns structure and appearance information in both the source and the target domains. It is noteworthy that even if both $y_t$ and $x^\prime$ belong to arbitrary identities, our effective disentangled module is capable of learning meaningful structure and appearance information of each identity.
During inference, we concatenate the appearance prior of $x^\prime$ and the structure prior of $y_t$ (small cubes in red and orange) in the latent space, and the reconstructed face $d_t$ shares the same structure with $y_t$ and keeps the appearance of $x^\prime$. Our framework allows concatenations of structure and appearance latent codes extracted from arbitrary identities in inference and permits \textit{many-to-many face reenactment}.

In summary, DF-VAE is a new conditional variational auto-encoder \cite{convae} with robustness and scalability. It conditions on two posteriors in different domains. In the disentangled module, the separated design of two encoders $E_\alpha$ and $E_\beta$, the explicit structure heatmap, and the unpaired data construction jointly force $E_\alpha$ to learn structure information and $E_\beta$ to learn appearance information.

%
%
%
%

\noindent
\textbf{Style matching and fusion.}
%
To fix the obvious style mismatch problems as shown in Figure \ref{fig:stype_mismatch}, we introduce a masked adaptive instance normalization (MAdaIN) module. We place a typical AdaIN \cite{adain} network after the reenacted face $d_t$. In the face swapping scenario, we only need to adjust the style of the face area and use the original background. Therefore, we use a mask $m_t$ to guide AdaIN \cite{adain} network to focus on style matching of the face area. To avoid boundary artifacts, we apply Gaussian Blur to $m_t$ and get the blurred mask $m_t^b$. 

\begin{figure}
   \begin{center}
       \includegraphics[width=\linewidth]{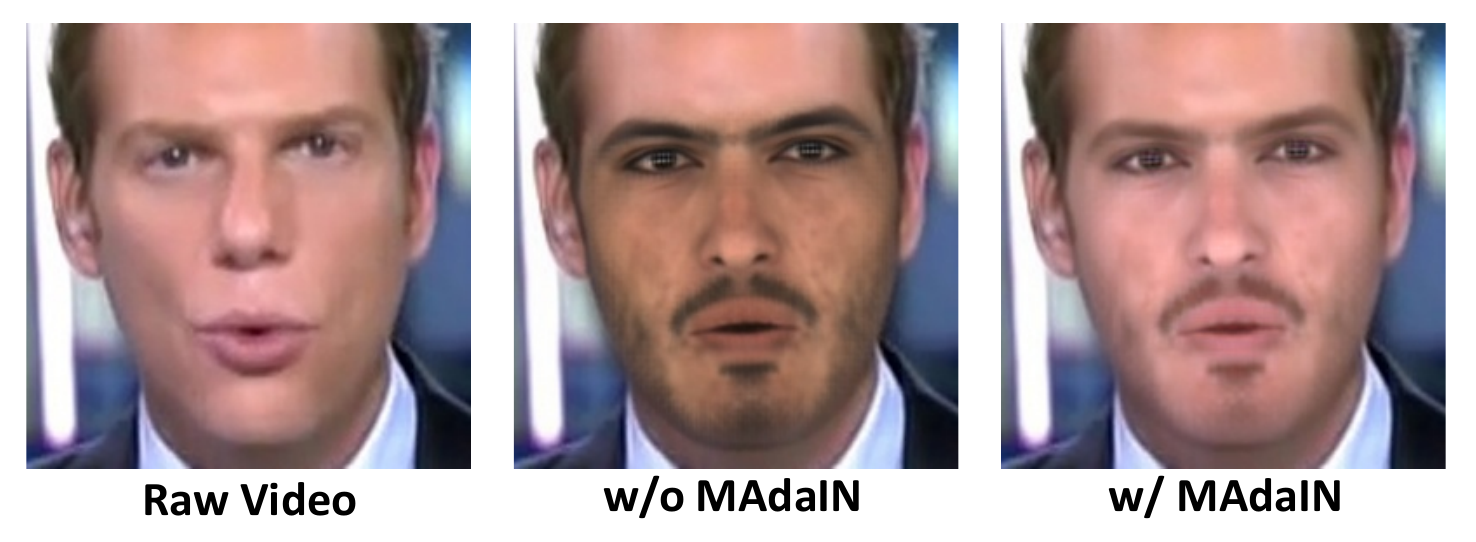}
   \end{center}
   \vspace{-0.35cm}
      \caption{Comparison of the swapped face styles without or with MAdaIN module.}
   \vspace{-0.45cm}
   \label{fig:style_matching}
\end{figure}

In our face swapping context, $d_t$ is the content input of MAdaIN, $y_t$ is the style input. MAdaIN adaptively computes the affine parameters from the face area of the style input:
\begin{equation}
\label{eq:8}
\begin{split}
    \rm{MAdaIN}\left(c, s\right)=\sigma\left(s\right)\left(\frac{c-\mu\left(c\right)}{\sigma\left(c\right)}\right)+\mu\left(s\right),
\end{split}
\end{equation}
where $c=m_t^b\cdot{d_t}$, $s=m_t^b\cdot{y_t}$. With the very low-cost MAdaIN module, we reconstruct $d_t$ again by Decoder $D_\delta$. The blurred mask $m_t^b$ is used again to fuse the reconstructed image with the background of $y_t$. At last, we get the swapped face $\overline{d}_t$. Figure~\ref{fig:style_matching} shows the effectiveness of MAdaIN module for style matching and fusion.

The MAdaIN module is jointly trained with the disentangled module in an end-to-end manner. Thus, by a \textit{single} model, DF-VAE can perform \textit{many-to-many face swapping} with obvious reduction of style mismatch and facial boundary artifacts (see Figure~\ref{fig:m2m} for the face swapping between three source identities and three target identities). Even if there are multiple identities in both the source domain and the target domain, the quality of face swapping does not degrade.


\noindent
\textbf{Temporal consistency constraint.} Temporal discontinuity of fake videos leads to obvious flickering of the face area, making them very easy to be spotted by forgery detection methods and human eyes. To improve temporal continuity, we let the disentangled module to learn temporal information of both the source face and the target face.

For simplification, we make a Markov assumption that the generation of the frame at time $t$ sequentially depends on its previous $P$ frames $\mathbf{x}_{(t-p):(t-1)}$. In our experiment, we set $P=1$ to balance quality improvement and training time.

In order to build the relationship between a current frame and previous ones, we further make an intuitive assumption that the optical flows should remain unchanged after reconstruction. We use FlowNet 2.0 \cite{flownet2} to estimate the optical flow $\tilde{x}_f$ \textit{w.r.t.} $\tilde{x}_t$ and $x_{t-1}$, $x_f$ \textit{w.r.t.} $x_t$ and $x_{t-1}$. Since face swapping is sensitive to minor facial details which can be greatly affected by flow estimation, we do not warp $x_{t-1}$ by the estimated flow like \cite{vid2vid}. Instead, we minimize the difference between $\tilde{x}_f$ and $x_f$ to improve temporal continuity while keeping stable facial detail generation. To this end, we propose a new temporal consistency constraint, which can be written as:
\begin{equation}
\label{eq:11}
\begin{split}
    L_{temporal}=\frac{1}{CHW}\|\tilde{x}_f-x_f\|_1,
\end{split}
\end{equation}
where $C=2$ for a common form of optical flow.

We only discuss the temporal continuity \textit{w.r.t.} the source face in this section because the case of the target face is the same. If multiple identities exist in one domain, temporal information of all these identities can be learned in an end-to-end manner.

\subsection{Scale and Diversity}

\begin{table}
\centering
\resizebox{0.8\linewidth}{!}{
\begin{tabular}{c|c}
\Xhline{1.2pt}
No.   &  Distortion Type  \\
\Xhline{1.2pt}
1 & Color saturation change  \\
2  & Local block-wise distortion \\
3 &  Color contrast change \\ 
4  & Gaussian blur \\
5 & White Gaussian noise in color components \\
6  & JPEG compression \\
7  & Video compression rate change  \\
\Xhline{1.2pt}
\end{tabular}
}
\caption{Seven types of distortions in DeeperForensics-$1.0$.}
\label{tab:distortions}
\vspace{-0.1cm}
\end{table}

\label{scaleanddiversity}
Our extensive data collection and the proposed DF-VAE method are designed to improve the \textit{quality} of manipulated videos in DeeperForensics-$1.0$ dataset. In this section, we will mainly discuss the \textit{scale} and \textit{diversity} aspects.

We provide $10,000$ manipulated videos with $5$ million frames. It is also an order of magnitude more than the previous datasets.
We take $1,000$ refined YouTube videos collected by FaceForensics++~\cite{FF++data} as the target videos. Each face of our collected $100$ identities is swapped onto $10$ target videos, thus $1,000$ raw manipulated videos are generated directly by DF-VAE in an end-to-end process.
Thanks to the scalability and multimodality of DF-VAE, the time overhead of model training and data generation is reduced to $1/5$ compared to the common Deepfakes methods, with no degradation in quality. Thus, a larger-scale dataset construction is possible.



To ensure \textit{diversity}, we apply various perturbations to better simulate videos in real scenes. Specifically, as shown in Table~\ref{tab:distortions}, seven types of distortions defined in Image Quality Assessment (IQA)~\cite{kwanyee2018hallucinated,TID2013} are included. Each of these distortions is divided into five intensity levels. We apply random-type distortions to the $1,000$ raw manipulated videos at five different intensity levels, producing a total of $5,000$ manipulated videos. Besides, an additional of $1,000$ robust manipulated videos are generated by adding random-type, random-level distortions to the $1,000$ raw manipulated videos.
Moreover, in contrast to all the previous datasets, each sample of another $3,000$ manipulated videos in DeeperForensics-$1.0$ is subjected to a mixture of more than one distortion. The variability of perturbations improves the \textit{diversity} of DeeperForensics-$1.0$ to better imitate the data distribution of real-world scenarios.

DeeperForensics-$1.0$ is a new \textit{large-scale} dataset consisting of over $60,000$ videos with $17.6$ million frames for real-world face forgery detection. \textit{High-quality} source videos and manipulated videos constitute two main contributions of the dataset. The \textit{diversity} of perturbations applying to the manipulated videos ensures the robustness of DeeperForensics-$1.0$ to simulate real scenes. The whole dataset is released, free to all research communities, for developing face forgery detection and more general human-face-related research.


\subsection{User Study}
To examine the quality of DeeperForensics-$1.0$ dataset, we engage $100$ professional participants, most of whom specialize in computer vision research. We believe these participants are qualified and well-trained in assessing realness of tempered videos. The user study is conducted on DeeperForensics-$1.0$ and six former datasets, \ie, UADFV \cite{UADFV}, DeepFake-TIMIT \cite{DFTIMIT}, Celeb-DF \cite{celebdf}, FaceForensics++ \cite{FF++data}, Deep Fake Detection \cite{google}, DFDC \cite{DFDC}. We randomly select $30$ video clips from each of these datasets and prepare a platform for the participants to evaluate their realness. Similar to the user study of \cite{deepvideo}, the participants are asked to provide their feedbacks to the statement ``The video clip looks real.'' and give scores at five levels ($1$-clearly disagree, $2$-weakly disagree, $3$-borderline, $4$-weakly agree, $5$-clearly agree. We assume that users who give a score of $4$ or $5$ think the video is ``real''). The user study results are presented in Table~\ref{tab:userstudy}. The quality of our dataset is appreciated by most of the participants. Compared to the previous datasets, DeeperForensics-$1.0$ achieves the highest realism rating. Although Celeb-DF \cite{celebdf} also gets very high realness scores, the scale of our dataset is much larger.

\begin{table}
\centering
\resizebox{1.0\linewidth}{!}{
\begin{tabular}{c|ccccc|c}
\Xhline{1.2pt}
Dataset     &  1  &  2  &  3  &  4  &  5  &  ``real'' \\
\Xhline{1.2pt}
UADFV \cite{UADFV} & 29.2 & 36.0 & 20.7 & 8.9 & 5.2 & 14.1\% \\
DeepFake-TIMIT \cite{DFTIMIT} & 31.4 & 31.4 & 24.8 & 9.6 & 2.7 & 12.3\% \\
Celeb-DF \cite{celebdf}  & 5.6 & 14.8 & 18.6 & 24.2 & 36.9 & 61.0\% \\
FaceForensics++ \cite{FF++data}  & 46.8  & 31.4 & 13.4 & 4.4 & 4.0 & 8.4\% \\
Deep Fake Detection \cite{google}  & 26.0 & 28.0 & 24.1 & 11.5 & 10.3 & 21.9\% \\
DFDC \cite{DFDC} & 25.4 & 29.7 & 22.0 & 11.9 & 11.1 & 23.0\% \\
\Xhline{1.2pt}
\textbf{DeeperForensics-1.0 (Ours)} & 4.3 & 8.9 & 22.6 & 29.8 & 34.3 & \textbf{64.1\%} \\
\Xhline{1.2pt}
\end{tabular}
}
\caption{The percentage of user study ratings for UADFV, DeepFake-TIMIT, Celeb-DF, FaceForensics++, Deep Fake Detection, DFDC, and DeeperForensics-$1.0$ dataset. A higher score means the users think the videos are more realistic.}
\label{tab:userstudy}
\vspace{-0.2cm}
\end{table}

\section{Video Forgery Detection Benchmark}

\label{benchmark}
\noindent
\textbf{Dataset split.}
In our benchmark, we exploit $1,000$ raw manipulated videos in Section~\ref{scaleanddiversity} and $1,000$ YouTube videos from FaceForensics++ \cite{FF++data} as our \textit{standard} set. The videos are split into training, validation, and test set with a ratio of $7:1:2$. The identities of the swapped faces may be duplicated because faces of $100$ invited actors are swapped onto $1,000$ driving videos. To avoid data leak, we randomly choose unrepeated $70$, $10$, and $20$ identities, and group all the videos according to the identities. Similar to~\cite{FF++data}, the test and training sets share a close distribution in our \textit{standard} set.

Other experiments in our benchmark are conducted on different variants of the standard set. These variants share the same $1,000$ driving videos with the standard set. We will detail them in Section~\ref{resultsandanalysis}. For a fair comparison, all the experiments are conducted in the same split setting.

\noindent
\textbf{Hidden test set.}
For real-world scenarios, some experiments conducted in previous works~\cite{celebdf,FF++data} may not perform a convincing evaluation due to the huge biases caused by a close distribution between the training and the test sets. The aforementioned standard set has the same setting with these works. As a result, strong detection baselines obtain very high accuracy on the standard test set as demonstrated in Section~\ref{resultsandanalysis}. However, the ultimate goal of the face forensics dataset is to help detect forgery in real scenes. Even if the accuracy on the standard test set is high, the models may easily fail in real-world scenarios. 

We argue that the test set of \textit{real-world} face forgery detection \textit{should not} share a close distribution with the training set. What we need is a test set that better simulates the real-world setting. We call it ``hidden'' test set. To better imitate fake videos in the real scene, the hidden test set should satisfy three factors:
1) \textit{Multiple sources.} Fake videos in-the-wild should be manipulated by different unknown methods.
2) \textit{High quality.} Threatening fake videos should have high quality to fool human eyes.
3) \textit{Diverse distortions.} Different perturbations should be taken into consideration.

Thus, in our initial benchmark, we introduce a challenging hidden test set with $400$ carefully selected videos. First, we collect fake videos generated by several unknown face swapping methods to ensure multiple sources. Then, we obscure all selected videos multiple times with diverse hidden distortions that are commonly seen in real scenes. Finally, we only select videos that can fool at least $50$ out of $100$ human observers in a user study. The ground truth labels are hidden and are used on our host server to evaluate the accuracy of detection models. Besides, the hidden test set will be enlarged constantly to get future versions along with development of Deepfakes technology. Fake videos manipulated by future face swapping methods will be included as long as they can pass the human test supported by us.


\subsection{Baselines}

Existing studies~\cite{celebdf,FF++data} primarily provide image-level face forgery detection benchmark. However, fake videos in-the-wild are much more menacing than manipulated images. We propose to conduct evaluation mainly based on video classification methods for two reasons. First, image-level face forgery detection methods do not consider any temporal information -- an important cue for video-based tasks. Second, image-level methods have been widely studied. We only choose one image-level method, XceptionNet \cite{xception}, which achieves the best performance in \cite{FF++data}, as one part of our benchmark for reference. The other four video-based baselines are C3D \cite{c3d}, TSN \cite{tsn}, I3D \cite{i3d}, and ResNet+LSTM \cite{resnet, lstm}, all of which have achieved promising results in video classification tasks. Details of all the baselines will be introduced in our \textit{Appendix}.

\subsection{Results and Analysis}
\label{resultsandanalysis}

Owing to the goal of detecting fakes in real-world scenarios, we mainly explore how common distortions appearing in real scenes affect the model performance. Accuracies of face forgery detection on the standard test set and the introduced hidden test set are evaluated under various settings. 



\begin{table}
\centering
\resizebox{1.0\linewidth}{!}{
\begin{tabular}{c|c|c|c|c|c}
\Xhline{1.2pt}
Train &  FF++ DF  & FF++ F2F &    FF++ FS & FF++ NT & \textbf{DeeperForensics-1.0} \\
Test (acc) &   hidden   & hidden &  hidden & hidden & \textbf{hidden} \\
\Xhline{1.2pt}
C3D \cite{c3d} & 57.50   & 57.75  &  52.13  & 58.25  & \textbf{74.75}  \\
TSN \cite{tsn} & 57.63   & 57.25  &  53.50  & 57.38  & \textbf{77.00} \\ 
I3D \cite{i3d} & 56.63   & 58.38  &  54.63  & 63.63  & \textbf{79.25} \\
ResNet+LSTM \cite{resnet, lstm}  & 57.38 & 56.13 & 54.88 & 59.50 & \textbf{78.25} \\
XceptionNet \cite{xception} & 57.38 & 58.75 & 54.75 & 57.38 & \textbf{77.00} \\
\Xhline{1.2pt}
\end{tabular}
}
\caption{The binary detection accuracy of the baselines on the hidden test set when trained on four manipulated methods in FaceForensics++ (FF++): DeepFakes (DF), Face2Face (F2F), FaceSwap (FS), NeuralTextures (NT), and on DeeperForensics-$1.0$ standard training set without distortions.}
\label{tab:setting1}
\end{table}

\noindent
\textbf{Evaluation of effectiveness of DeeperForensics-1.0.} For a fair comparison, we evaluate DeeperForensics-$1.0$ and the state-of-the-art FaceForensics++ \cite{FF++data} dataset because they use the same driving videos. In this setting, we use $1,000$ raw manipulated videos without distortions in the standard set of DeeperForensics-$1.0$. For FaceForensics++, the same split is applied to its \textit{four} subsets. All the models are tested on the hidden test set (see Table~\ref{tab:setting1}).

The baselines trained on the standard training set of DeeperForensics-$1.0$ achieve much better performance on the hidden test set than all the \textit{four} subsets of FaceForensics++. This proves the higher \textit{quality} of DeeperForensics-$1.0$ over prior works, making it more useful for real-world face forgery detection.
In Table~\ref{tab:setting1}, I3D \cite{i3d} obtains the best performance on the hidden test set when trained on the standard training set. We conjecture that the temporal discontinuity of fake videos leads to higher accuracy by this video-level forgery detection method.



\begin{table}
\centering
\resizebox{1.0\linewidth}{!}{
\begin{tabular}{c|c|cc|cc|cc}
\Xhline{1.2pt}
Train       &  \textbf{std}   & std  & std & std/sing & std/rand & std/sing & std/rand \\
Test (acc)    & \textbf{std}  & std/sing & std/rand & std/sing & std/rand & std/rand & std/ sing\\
\Xhline{1.2pt}
C3D \cite{c3d} & \textbf{98.50} & 87.63 & 92.38 & 95.38 & 96.63 & 96.75 & 94.00 \\
TSN \cite{tsn} & \textbf{99.25} & 91.50 & 95.00 & 98.25 & 98.88 & 98.12 & 99.12 \\ 
I3D \cite{i3d} & \textbf{100.00} & 90.75 & 96.88 & 99.50 & 99.63 & 99.63 & 98.00 \\
ResNet+LSTM \cite{resnet, lstm} & \textbf{100.00} & 90.63 & 97.13 & 100.00& 98.63 & 100.00& 97.25 \\
XceptionNet \cite{xception} & \textbf{100.00} & 88.38 & 94.75 & 99.63 & 99.63 & 99.75 & 99.00 \\
\Xhline{1.2pt}
\end{tabular}
}
\caption{The binary detection accuracy of the baselines when trained and tested on DeeperForensics-$1.0$ dataset with different distortion perturbations. We analyze different training and testing settings on the standard set without distortions (std), the standard set with single-level distortions (std/sing), and the standard set with random-level distortions (std/rand).}
\label{tab:setting2}
\end{table}

\noindent
\textbf{Evaluation of dataset perturbations.} We study the effect of perturbations towards the forgery detection model performance. In contrast to prior work \cite{FF++data}, we try to evaluate the baseline accuracies when applying different distortions to the training and the test sets, in order to explore the function of perturbations in face forensics dataset.


In this setting, we conduct all the experiments on DeeperForensics-$1.0$ dataset with high diversity of perturbations. We use $1,000$ manipulated videos in the standard set (std), $1,000$ manipulated videos with single-level (level-5), random-type distortions (std/sing), $1,000$ manipulated videos with random-level, random-type distortions (std/rand). The data split is the same as that of the standard set with a ratio of $7:1:2$.

In Column $2$ of Table~\ref{tab:setting2}, we find the accuracy is nearly $100\%$ when the models are trained and tested on the standard set. This is reasonable because the strong baselines perform very well in a clean dataset with the same distribution.
In Columns $3$ and $4$, the accuracy decrease compared to Column $2$, when we choose std/sing and std/rand as the test set. Most of the video-level methods except C3D \cite{c3d} are more robust to perturbations on test set than XceptionNet \cite{xception}. This setting is very common because different distributions of the training and the test sets lead to decrease in model accuracies. Hence, the lack of perturbations in the face forensics dataset cutbacks the model performance for real-world face forgery detection with even more complex data distribution.

When we apply corresponding distortions to the training and test sets, the accuracy will increase (Column $5$ and $6$ in Table~\ref{tab:setting2}) compared to Column $3$ and $4$. However, this setting is impractical because the distributions of the training and test sets are still the same. We should augment the test set to better simulate the real-world distribution. Thus, some evaluation settings in previous works \cite{celebdf,FF++data} are unreasonable. If we swap the training set and the test set of std/sing and std/rand to further randomize the condition, results shown in Column $7$ and $8$ indicate that the accuracy remains high. This evaluation setting shows the possibility that with the same generation method, exerting appropriate distortions to the training set can make face forgery detection models more robust to real-world perturbations.

\begin{table}
\centering
\resizebox{0.95\linewidth}{!}{
\begin{tabular}{c|c|c|c|c}
\Xhline{1.2pt}
Train        & std  & std+std/sing & std+std/rand & \textbf{std+std/mix}  \\
Test (acc)     & hidden & hidden & hidden & \textbf{hidden}  \\
\Xhline{1.2pt}
C3D \cite{c3d}  & 74.75  & 78.25  & 78.13 &  \textbf{78.88}   \\
TSN \cite{tsn}  & 77.00  & 78.75  & 79.50 & \textbf{79.50} \\ 
I3D \cite{i3d}  & 79.25  & 80.13  & 80.13 & \textbf{80.13} \\
ResNet+LSTM \cite{resnet, lstm}  & 78.25 & 80.25 & 79.50 &  \textbf{80.25} \\
XceptionNet \cite{xception} & 77.00 & 79.75 & 79.75 &  \textbf{79.88} \\
\Xhline{1.2pt}
\end{tabular}
}

\caption{The binary detection accuracy of the baselines on the hidden test set when trained on DeeperForensics-$1.0$ dataset with the standard set without distortions (std), combination of std and the standard set with single-level distortions (std+std/sing), combination of std and the standard set with random-level distortions (std+std/rand), combination of std and the standard set with the mixed distortions(std+std/mix).}
\label{tab:setting3}
\end{table}

\noindent
\textbf{Evaluation of variants of training set for real-world face forgery detection.} We have conducted several experiments for evaluations of possible perturbations. Nevertheless, the case is more complex in real scenes because no information about the fake videos is available. The video may be subjected to more than one type and diverse levels of distortions. In addition to distortions, the method manipulating the faces is unknown.

From the evaluation of perturbations, we find the possibility of augmenting the training set to improve detection model performance. Thus, we further evaluate baseline performance on the hidden test set by devising some variants of the training set.
We perform experiments on DeeperForensics-$1.0$. In this setting, other than std, std/sing, and std/rand, we use additional $1,000$ manipulated videos, each of which is subjected to a mixture of three random-level, random-type distortions (std/mix). We combine std with std/sing, std/rand, and std/mix, respectively, yielding three new training sets (with the same data split as the former settings).

Column $2$ in Table~\ref{tab:setting3} shows the low accuracy when the models trained on std and tested on the hidden test set (same as Column $6$ in Table~\ref{tab:setting1}). Columns $3$ and $4$ indicate that the accuracy of all the baseline models increase when trained on std+std/sing and std+std/rand. The accuracy of I3D \cite{i3d} and ResNet+LSTM \cite{resnet, lstm}, are over $80\%$ in some cases. In a more complex setting, when the models are trained on std+std/mix, Column $5$ shows the accuracy of all the detection baselines further increase.

The results suggest that designing suitable training set variants has the potential to help increase face forgery detection accuracy, and applying various distortions to ensure the \textit{diversity} of DeeperForensics-$1.0$ is necessary. In addition, compared to image-level method, video-level face forgery detection methods have more potential capabilities to crack real-world fake videos as shown in Table~\ref{tab:setting3}.

Although the accuracy on the challenging hidden test set is still not very high, we provide two initial directions for future real-world face forgery detection research:
1) Improving the source data collection and generation method to ensure the \textit{quality} of the training set;
2) Augmenting the training set by various distortions to ensure its \textit{diversity}. We welcome researchers to make our benchmark more comprehensive.

\section{Discussion}

In this work, we propose a new large-scale dataset named DeeperForensics-$1.0$ to facilitate the research of face forgery detection towards \textit{real-world} scenarios. We make several efforts to ensure \textit{good quality}, \textit{large scale}, and \textit{high diversity} of this dataset. Based on the dataset, we further benchmark existing representative forgery detection methods, offering insights into the current status and future strategy in face forgery detection.
Several topics can be considered as future works.
1) We will continue to collect more source and target videos to further expand DeeperForensics.
2) We plan to invite interested researchers for contributing their video falsification methods to enlarge our hidden test set, as long as the fakes can pass the human test supported by us.
3) A better evaluation metric for face forgery detection methods is also an interesting research topic.
\\[5pt]
\noindent
\textbf{Acknowledgments.} This work is supported by the SenseTime-NTU Collaboration Project, Singapore MOE AcRF Tier 1 (2018-T1-002-056), NTU SUG, and NTU NAP. We gratefully acknowledge the exceptional help from Hao Zhu and Keqiang Sun for their contribution on source data collection and coordination.

{\small
\balance
\bibliographystyle{ieee_fullname}
\bibliography{egbib}

\begin{thebibliography}{10}\itemsep=-1pt

\bibitem{DFL}
Deepfacelab.
\newblock \url{https://github.com/iperov/DeepFaceLab/}.
\newblock Accessed: 2019-08-20.

\bibitem{DeepFakes}
Deepfakes.
\newblock \url{https://github.com/deepfakes/faceswap/}.
\newblock Accessed: 2019-08-16.

\bibitem{FaceSwap}
Faceswap.
\newblock \url{https://github.com/MarekKowalski/FaceSwap/}.
\newblock Accessed: 2019-08-18.

\bibitem{faceswap-GAN}
faceswap-gan.
\newblock \url{https://github.com/shaoanlu/faceswap-GAN/}.
\newblock Accessed: 2019-08-16.

\bibitem{fakeapp}
Fakeapp.
\newblock \url{https://www.fakeapp.com/}.
\newblock Accessed: 2019-07-25.

\bibitem{mosonet}
Darius Afchar, Vincent Nozick, Junichi Yamagishi, and Isao Echizen.
\newblock Mesonet: a compact facial video forgery detection network.
\newblock In {\em WIFS}, 2018.

\bibitem{miccf2000}
Irene Amerini, Lamberto Ballan, Roberto Caldelli, Alberto Del~Bimbo, and
  Giuseppe Serra.
\newblock A sift-based forensic method for copy--move attack detection and
  transformation recovery.
\newblock {\em TIFS}, 6:1099--1110, 2011.

\bibitem{FF++bm3}
Belhassen Bayar and Matthew~C Stamm.
\newblock A deep learning approach to universal image manipulation detection
  using a new convolutional layer.
\newblock In {\em IH \& MMSEC}, 2016.

\bibitem{google}
Google~AI Blog.
\newblock Contributing data to deepfake detection research.
\newblock
  \url{https://ai.googleblog.com/2019/09/contributing-data-to-deepfake-detection.html}.
\newblock Accessed: 2019-09-25.

\bibitem{3dmm}
Chen Cao, Yanlin Weng, Shun Zhou, Yiying Tong, and Kun Zhou.
\newblock Facewarehouse: A 3d facial expression database for visual computing.
\newblock {\em TVCG}, 20:413--425, 2013.

\bibitem{i3d}
Joao Carreira and Andrew Zisserman.
\newblock Quo vadis, action recognition? a new model and the kinetics dataset.
\newblock In {\em CVPR}, 2017.

\bibitem{xception}
Fran{\c{c}}ois Chollet.
\newblock Xception: Deep learning with depthwise separable convolutions.
\newblock In {\em CVPR}, 2017.

\bibitem{FF++bm2}
Davide Cozzolino, Giovanni Poggi, and Luisa Verdoliva.
\newblock Recasting residual-based local descriptors as convolutional neural
  networks: an application to image forgery detection.
\newblock In {\em IH \& MMSEC}, 2017.

\bibitem{DFDC}
Brian Dolhansky, Russ Howes, Ben Pflaum, Nicole Baram, and Cristian~Canton
  Ferrer.
\newblock The deepfake detection challenge (dfdc) preview dataset.
\newblock {\em arXiv preprint}, arXiv:1910.08854, 2019.

\bibitem{FF++bm1}
Jessica Fridrich and Jan Kodovsky.
\newblock Rich models for steganalysis of digital images.
\newblock {\em TIFS}, 7:868--882, 2012.

\bibitem{gan}
Ian Goodfellow, Jean Pouget-Abadie, Mehdi Mirza, Bing Xu, David Warde-Farley,
  Sherjil Ozair, Aaron Courville, and Yoshua Bengio.
\newblock Generative adversarial nets.
\newblock In {\em NeurIPS}, 2014.

\bibitem{resnet}
Kaiming He, Xiangyu Zhang, Shaoqing Ren, and Jian Sun.
\newblock Deep residual learning for image recognition.
\newblock In {\em CVPR}, 2016.

\bibitem{fid}
Martin Heusel, Hubert Ramsauer, Thomas Unterthiner, Bernhard Nessler, and Sepp
  Hochreiter.
\newblock Gans trained by a two time-scale update rule converge to a local nash
  equilibrium.
\newblock In {\em NeurIPS}, 2017.

\bibitem{lstm}
Sepp Hochreiter and J{\"u}rgen Schmidhuber.
\newblock Long short-term memory.
\newblock {\em Neural computation}, 9:1735--1780, 1997.

\bibitem{adain}
Xun Huang and Serge Belongie.
\newblock Arbitrary style transfer in real-time with adaptive instance
  normalization.
\newblock In {\em ICCV}, 2017.

\bibitem{flownet2}
Eddy Ilg, Nikolaus Mayer, Tonmoy Saikia, Margret Keuper, Alexey Dosovitskiy,
  and Thomas Brox.
\newblock Flownet 2.0: Evolution of optical flow estimation with deep networks.
\newblock In {\em CVPR}, 2017.

\bibitem{BN}
Sergey Ioffe and Christian Szegedy.
\newblock Batch normalization: Accelerating deep network training by reducing
  internal covariate shift.
\newblock In {\em ICML}, 2015.

\bibitem{deepvideo}
Hyeongwoo Kim, Pablo Carrido, Ayush Tewari, Weipeng Xu, Justus Thies, Matthias
  Niessner, Patrick P{\'e}rez, Christian Richardt, Michael Zollh{\"o}fer, and
  Christian Theobalt.
\newblock Deep video portraits.
\newblock {\em ACM TOG}, 37:163, 2018.

\bibitem{adam}
Diederik~P Kingma and Jimmy Ba.
\newblock Adam: A method for stochastic optimization.
\newblock {\em arXiv preprint}, arXiv:1412.6980, 2014.

\bibitem{convae}
Durk~P Kingma, Shakir Mohamed, Danilo~Jimenez Rezende, and Max Welling.
\newblock Semi-supervised learning with deep generative models.
\newblock In {\em NeurIPS}, 2014.

\bibitem{vae}
Diederik~P Kingma and Max Welling.
\newblock Auto-encoding variational bayes.
\newblock {\em arXiv preprint}, arXiv:1312.6114, 2013.

\bibitem{DFTIMIT}
Pavel Korshunov and S{\'e}bastien Marcel.
\newblock Deepfakes: a new threat to face recognition? assessment and
  detection.
\newblock {\em arXiv preprint}, arXiv:1812.08685, 2018.

\bibitem{rtddataset}
Pawe{\l} Korus and Jiwu Huang.
\newblock Multi-scale analysis strategies in prnu-based tampering localization.
\newblock {\em TIFS}, 12:809--824, 2016.

\bibitem{cdfbm2}
Yuezun Li and Siwei Lyu.
\newblock Exposing deepfake videos by detecting face warping artifacts.
\newblock {\em arXiv preprint}, arXiv:1811.00656, 2018.

\bibitem{celebdf}
Yuezun Li, Xin Yang, Pu Sun, Honggang Qi, and Siwei Lyu.
\newblock Celeb-df: A new dataset for deepfake forensics.
\newblock {\em arXiv preprint}, 2019.

\bibitem{kwanyee2018hallucinated}
Kwan-Yee Lin and Guangxiang Wang.
\newblock Hallucinated-iqa: No-reference image quality assessment via
  adversarial learning.
\newblock In {\em CVPR}, 2018.

\bibitem{cdfbm3}
Falko Matern, Christian Riess, and Marc Stamminger.
\newblock Exploiting visual artifacts to expose deepfakes and face
  manipulations.
\newblock In {\em WACVW}, 2019.

\bibitem{stackedhourglass}
Alejandro Newell, Kaiyu Yang, and Jia Deng.
\newblock Stacked hourglass networks for human pose estimation.
\newblock In {\em ECCV}, 2016.

\bibitem{cdfbm4}
Huy~H Nguyen, Fuming Fang, Junichi Yamagishi, and Isao Echizen.
\newblock Multi-task learning for detecting and segmenting manipulated facial
  images and videos.
\newblock {\em arXiv preprint}, arXiv:1906.06876, 2019.

\bibitem{TID2013}
Nikolay Ponomarenko, Lina Jin, Oleg Ieremeiev, Vladimir Lukin, Karen
  Egiazarian, Jaakko Astola, Benoit Vozel, Kacem Chehdi, Marco Carli, Federica
  Battisti, et~al.
\newblock Image database tid2013: Peculiarities, results and perspectives.
\newblock {\em Signal Processing: Image Communication}, 30:57--77, 2015.

\bibitem{FF++bm4}
Nicolas Rahmouni, Vincent Nozick, Junichi Yamagishi, and Isao Echizen.
\newblock Distinguishing computer graphics from natural images using
  convolution neural networks.
\newblock In {\em WIFS}, 2017.

\bibitem{FF++data}
Andreas R{\"o}ssler, Davide Cozzolino, Luisa Verdoliva, Christian Riess, Justus
  Thies, and Matthias Nie{\ss}ner.
\newblock Faceforensics++: Learning to detect manipulated facial images.
\newblock {\em arXiv preprint}, arXiv:1901.08971, 2019.

\bibitem{is}
Tim Salimans, Ian Goodfellow, Wojciech Zaremba, Vicki Cheung, Alec Radford, and
  Xi Chen.
\newblock Improved techniques for training gans.
\newblock In {\em NeurIPS}, 2016.

\bibitem{VIDTIMIT}
Conrad Sanderson.
\newblock The vidtimit database.
\newblock Technical report, IDIAP, 2002.

\bibitem{vgg}
Karen Simonyan and Andrew Zisserman.
\newblock Very deep convolutional networks for large-scale image recognition.
\newblock {\em arXiv preprint}, arXiv:1409.1556, 2014.

\bibitem{NeuralTextures}
Justus Thies, Michael Zollh{\"o}fer, and Matthias Nie{\ss}ner.
\newblock Deferred neural rendering: Image synthesis using neural textures.
\newblock {\em arXiv preprint}, arXiv:1904.12356, 2019.

\bibitem{face2face}
Justus Thies, Michael Zollhofer, Marc Stamminger, Christian Theobalt, and
  Matthias Nie{\ss}ner.
\newblock Face2face: Real-time face capture and reenactment of rgb videos.
\newblock In {\em CVPR}, 2016.

\bibitem{c3d}
Du Tran, Lubomir Bourdev, Rob Fergus, Lorenzo Torresani, and Manohar Paluri.
\newblock Learning spatiotemporal features with 3d convolutional networks.
\newblock In {\em ICCV}, 2015.

\bibitem{IN}
Dmitry Ulyanov, Andrea Vedaldi, and Victor Lempitsky.
\newblock Instance normalization: The missing ingredient for fast stylization.
\newblock {\em arXiv preprint}, arXiv:1607.08022, 2016.

\bibitem{tsn}
Limin Wang, Yuanjun Xiong, Zhe Wang, Yu Qiao, Dahua Lin, Xiaoou Tang, and Luc
  Van~Gool.
\newblock Temporal segment networks: Towards good practices for deep action
  recognition.
\newblock In {\em ECCV}, 2016.

\bibitem{vid2vid}
Ting-Chun Wang, Ming-Yu Liu, Jun-Yan Zhu, Guilin Liu, Andrew Tao, Jan Kautz,
  and Bryan Catanzaro.
\newblock Video-to-video synthesis.
\newblock {\em arXiv preprint}, arXiv:1808.06601, 2018.

\bibitem{wayne2018lab}
Wayne Wu, Chen Qian, Shuo Yang, Quan Wang, Yici Cai, and Qiang Zhou.
\newblock Look at boundary: A boundary-aware face alignment algorithm.
\newblock In {\em CVPR}, 2018.

\bibitem{reenactgan}
Wayne Wu, Yunxuan Zhang, Cheng Li, Chen Qian, and Chen Change~Loy.
\newblock Reenactgan: Learning to reenact faces via boundary transfer.
\newblock In {\em ECCV}, 2018.

\bibitem{UADFV}
Xin Yang, Yuezun Li, and Siwei Lyu.
\newblock Exposing deep fakes using inconsistent head poses.
\newblock In {\em ICASSP}, pages 8261--8265, 2019.

\bibitem{wildweb}
Markos Zampoglou, Symeon Papadopoulos, and Yiannis Kompatsiaris.
\newblock Detecting image splicing in the wild (web).
\newblock In {\em ICMEW}, 2015.

\bibitem{cdfbm1}
Peng Zhou, Xintong Han, Vlad~I Morariu, and Larry~S Davis.
\newblock Two-stream neural networks for tampered face detection.
\newblock In {\em CVPRW}, 2017.

\end{thebibliography}
}

\clearpage
\section*{Appendix}
\label{appendix}
\appendix

\section{Derivation}
The core equations of DF-VAE are Eq.~\eqref{eq:4}, Eq.~\eqref{eq:5}, and Eq.~\eqref{eq:6}. We will provide the detailed mathematical derivations in this section.
\\[5pt]
\noindent
\textbf{Derivation of Eq.~\eqref{eq:4} and Eq.~\eqref{eq:5}:}
\begin{equation}
\tiny
\notag
\begin{split}
    &\quad\log{p_\theta\left(x_t|\hat{x}_t,x^\prime\right)}\\
    &=\mathbb{E}_{q_\phi\left(z_x|x_t,\hat{x}_t,x^\prime\right)}\left(\log{p_\theta\left(x_t|\hat{x}_t,x^\prime\right)}\right)\\
    &=\mathbb{E}_{q_\phi\left(z_x|x_t,\hat{x}_t,x^\prime\right)}\left[\log{\frac{p_\theta\left(x_t,z_x|\hat{x}_t,x^\prime\right)}{p_\theta\left(z_x|x_t,\hat{x}_t,x^\prime\right)}}\right]\\
    &=\mathbb{E}_{q_\phi\left(z_x|x_t,\hat{x}_t,x^\prime\right)}\left[\log{\frac{q_\phi\left(z_x|x_t,\hat{x}_t,x^\prime\right)}{p_\theta\left(z_x|x_t,\hat{x}_t,x^\prime\right)}}\cdot\frac{p_\theta\left(x_t,z_x|\hat{x}_t,x^\prime\right)}{q_\phi\left(z_x|x_t,\hat{x}_t,x^\prime\right)}\right]\\
    &=\int{q_\phi\left(z_x|x_t,\hat{x}_t,x^\prime\right)\left[\log{\frac{q_\phi\left(z_x|x_t,\hat{x}_t,x^\prime\right)}{p_\theta\left(z_x|x_t,\hat{x}_t,x^\prime\right)}}+\log{\frac{p_\theta\left(x_t,z_x|\hat{x}_t,x^\prime\right)}{q_\phi\left(z_x|x_t,\hat{x}_t,x^\prime\right)}}\right]dz_x}\\
    &=D_{KL}\left(q_\phi\left(z_x|x_t,\hat{x}_t,x^\prime\right)\|p_\theta\left(z_x|x_t,\hat{x}_t,x^\prime\right)\right)+L\left(\theta,\phi;x_t,\hat{x}_t,x^\prime\right), \\
\end{split}
\end{equation}
where
\begin{equation}
\scriptsize
\notag
\begin{split}
	&L\left(\theta,\phi;x_t,\hat{x}_t,x^\prime\right)=\mathbb{E}_{q_\phi\left(z_x|x_t,\hat{x}_t,x^\prime\right)}\left[\log{\frac{p_\theta\left(x_t,z_x|\hat{x}_t,x^\prime\right)}{q_\phi\left(z_x|x_t,\hat{x}_t,x^\prime\right)}}\right],\\
    &D_{KL}\left(q_\phi\left(z_x|x_t,\hat{x}_t,x^\prime\right)\|p_\theta\left(z_x|x_t,\hat{x}_t,x^\prime\right)\right)\geq0. \\
\end{split}
\end{equation}

\noindent
\textbf{Derivation of Eq.~\eqref{eq:6}:}
\begin{equation}
\tiny
\notag
\begin{split}
	&\quad L\left(\theta,\phi;x_t,\hat{x}_t,x^\prime\right)\\
	&=\mathbb{E}_{q_\phi\left(z_x|x_t,\hat{x}_t,x^\prime\right)}\left[\log{\frac{p_\theta\left(x_t,z_x|\hat{x}_t,x^\prime\right)}{q_\phi\left(z_x|x_t,\hat{x}_t,x^\prime\right)}}\right]\\
	&=\mathbb{E}_{q_\phi\left(z_x|x_t,\hat{x}_t,x^\prime\right)}
	\left[\log{
	\frac{p_\theta\left(x_t|z_x,\hat{x}_t,x^\prime\right)
	p_\theta\left(z_x|\hat{x}_t,x^\prime\right)}
	{q_\phi\left(z_x|x_t,\hat{x}_t,x^\prime\right)}}
	\right]\\
	&=\int{q_\phi\left(z_x|x_t,\hat{x}_t,x^\prime\right)
	\left[-\log{\frac{q_\phi\left(z_x|x_t,\hat{x}_t,x^\prime\right)}
	{p_\theta\left(z_x|\hat{x}_t,x^\prime\right)}}
	+\log{p_\theta\left(x_t|z_x,\hat{x}_t,x^\prime\right)}
	\right]dz_x} \\
	&=-D_{KL}\left(q_\phi\left(z_x|x_t,\hat{x}_t,x^\prime\right)\|p_\theta\left(z_x|\hat{x}_t,x^\prime\right)\right)\\
    &\qquad+\mathbb{E}_{q_\phi\left(z_x|x_t,\hat{x}_t,x^\prime\right)}\left[\log{p_\theta\left(x_t|z_x,\hat{x}_t,x^\prime\right)}\right].
\end{split}
\end{equation}

\section{Objective}
\noindent
\textbf{Reconstruction loss.} In the reconstruction, the source face and target face share the same forms of loss functions. The reconstruction loss of the source face, $L_{recon_x}$, can be written as:

\begin{equation}
\label{eq:12}
\begin{split}
    L_{recon_x}=\lambda_{r_1}L_{pixel}\left(\tilde{x},x\right)+\lambda_{r_2}L_{ssim}\left(\tilde{x},x\right).
\end{split}
\end{equation}

$L_{pixel}$ indicates pixel loss. It calculates the Mean Absolute Error (MAE) after reconstruction, which can be written as:
\begin{equation}
\label{eq:13}
\begin{split}
    L_{pixel}\left(\tilde{x},x\right)=\frac{1}{CHW}\|\tilde{x}-x\|_1.
\end{split}
\end{equation}

$L_{ssim}$ denotes ssim loss. It computes the Structural Similarity (SSIM) of the reconstructed face and the original face, which has the form of:
\begin{equation}
\label{eq:14}
\begin{split}
    L_{ssim}\left(\tilde{x},x\right)=\frac{\left(2\mu_{\tilde{x}}\mu_x+C_1\right)\left(2\sigma_{\tilde{x}x}+C_2\right)}{\left(\mu_{\tilde{x}}^2+\mu_x^2+C_1\right)\left(\sigma_{\tilde{x}}^2+\sigma_x^2+C_2\right)}.
\end{split}
\end{equation}

$\lambda_{r_1}$ and $\lambda_{r_2}$ are two hyperparameters that control the weights of two parts of the reconstruction loss. For the target face, we have the similar form of reconstruction loss:
\begin{equation}
\label{eq:15}
\begin{split}
    L_{recon_y}=\lambda_{r_1}L_{pixel}\left(\tilde{y},y\right)+\lambda_{r_2}L_{ssim}\left(\tilde{y},y\right).
\end{split}
\end{equation}

Thus, the full reconstruction loss can be written as:
\begin{equation}
\label{eq:16}
\begin{split}
    L_{recon}=L_{recon_x}+L_{recon_y}.
\end{split}
\end{equation}

\noindent
\textbf{KL loss.} Since DF-VAE is a new conditional variational auto-encoder, reparameterization trick is utilized to make the sampling operation differentiable by an auxiliary variable with independent marginal. We use the typical KL loss in \cite{vae} with the form of:
\begin{equation}
\small
\label{eq:17}
\begin{split}
    L_{KL}\left(q_\phi\left(z\right),p_\theta\left(z\right)\right)=\frac{1}{2}\sum_{j=1}^J{\left(1+\log\left(\left(\sigma_j\right)^2\right)-\left(\mu_j\right)^2-\left(\sigma_j\right)^2\right)},
\end{split}
\end{equation}
where $J$ is the dimensionality of the latent prior $z$, $\mu_j$ and $\sigma_j$ are the $j$-th element of variational mean and s.d. vectors, respectively.

\noindent
\textbf{MAdaIN loss.} The MAdaIN module is jointly trained with the disentangled module in an end-to-end manner. We apply MAdaIN loss for this module, in a similar form as described in \cite{adain}. We use the VGG-19 \cite{vgg} to compute MAdaIN loss to train Decoder $D_\delta$:
\begin{equation}
\small
\label{eq:18}
\begin{split}
    L_{\rm{MAdaIN}}=L_c+\lambda_{ma}L_s.
\end{split}
\end{equation}

$L_c$ denotes the content loss, which is the Euclidean distance between the target features and the features of the swapped face. $L_c$ has the form of:
\begin{equation}
\small
\label{eq:19}
\begin{split}
    L_c=\|o-c\|_2,
\end{split}
\end{equation}
where $o=m_t^b\cdot{\overline{d}_t}$, $c=m_t^b\cdot{d_t}$. $m_t^b$ is the blurred mask described in Section~\ref{DFVAE}.

$L_s$ represents the style loss, which matches the mean and standard deviation of the style features. Like \cite{adain}, we match the IN \cite{IN} statistics instead of using Gram matrix loss which can produce similar results. $L_s$ can be written as:
\begin{equation}
\small
\label{eq:20}
\begin{split}
    L_s=\sum_{i=1}^L\|\mu(\Phi_i(o))-\mu(\Phi_i(s))\|_2\\+\sum_{i=1}^L\|\sigma(\Phi_i(o))-\sigma(\Phi_i(s))\|_2,
\end{split}
\end{equation}
where $o=m_t^b\cdot{\overline{d}_t}$, $s=m_t^b\cdot{y_t}$. $m_t^b$ is the blurred mask described in Section~\ref{DFVAE}. $\Phi_i$ denotes the layer used in VGG-19 \cite{vgg}. Similar to \cite{adain}, we use \texttt{relu1\_1}, \texttt{relu2\_1}, \texttt{relu3\_1}, \texttt{relu4\_1} layers with equal weights.

$\lambda_{ma}$ is the weight of style loss to balance two parts of MAdaIN loss.

\noindent
\textbf{Temporal loss.} We have given a detailed introduction of temporal consistency constraint in Section~\ref{DFVAE}. The temporal loss has the form of Eq.~\eqref{eq:11}. We will not repeat it here.

\noindent
\textbf{Total objective.} DF-VAE is an end-to-end many-to-many face swapping framework. We jointly train all parts of the networks. The problem can be described as the optimization of the following total objective:
\begin{equation}
\small
\label{eq:21}
\begin{split}
    L_{total}=\lambda_1L_{recon}+\lambda_2L_{KL}+\lambda_3L_{MAdaIN}+\lambda_4L_{temporal},
\end{split}
\end{equation}
where $\lambda_1$, $\lambda_2$, $\lambda_3$, $\lambda_4$ are the weight hyperparameters of four types of loss functions introduced above.

\section{Implementation Details}
The whole DF-VAE framework is end-to-end. We use the pretrained stacked hourglass networks \cite{stackedhourglass} to extract landmarks. The numbers of stacks and blocks are set to $4$ and $1$, respectively. We exploit FlowNet 2.0 network \cite{flownet2} to estimate optical flows. The typical AdaIN network \cite{adain} is applied to our style matching and fusion module. The learning rate is set to $0.00005$ for all parts of DF-VAE. We utilize Adam \cite{adam} and set $\beta_1=0.5$, $\beta_2=0.999$. All the experiments are conducted on NVIDIA Tesla V100 GPUs.

\section{User Study of Methods}

In addition to user study based on datasets to examine the quality of DeeperForensics-$1.0$ dataset, we also carry out a user study to compare DF-VAE with state-of-the-art face manipulation methods. We will present the user study of methods in this section.

\noindent
\textbf{Baselines.} We choose three learning-based open-source methods as our baselines: DeepFakes \cite{DeepFakes}, faceswap-GAN \cite{faceswap-GAN}, and ReenactGAN \cite{reenactgan}. These three methods are representative, which are based on different architectures. DeepFakes \cite{DeepFakes} is a well-known method based on Auto-Encoders (AE). It uses a shared encoder and two separated decoders to perform face swapping. faceswap-GAN \cite{faceswap-GAN} is based on Generative Adversarial Networks (GAN) \cite{gan}, which has a similar structure as DeepFakes \cite{DeepFakes} but also uses a paired discriminators to improve face swapping quality. ReenactGAN \cite{reenactgan} makes a boundary latent space assumption and uses a transformer to adapt the boundary of source face to that of target face. As a result, ReenactGAN can perform many-to-one face reenactment. After getting the reenacted faces, we use our carefully designed fusion method to obtain the swapped faces. For a fair comparison, DF-VAE utilizes the same fusion method when compared to ReenactGAN \cite{reenactgan}.

\begin{figure}
   \begin{center}
       \includegraphics[width=\linewidth]{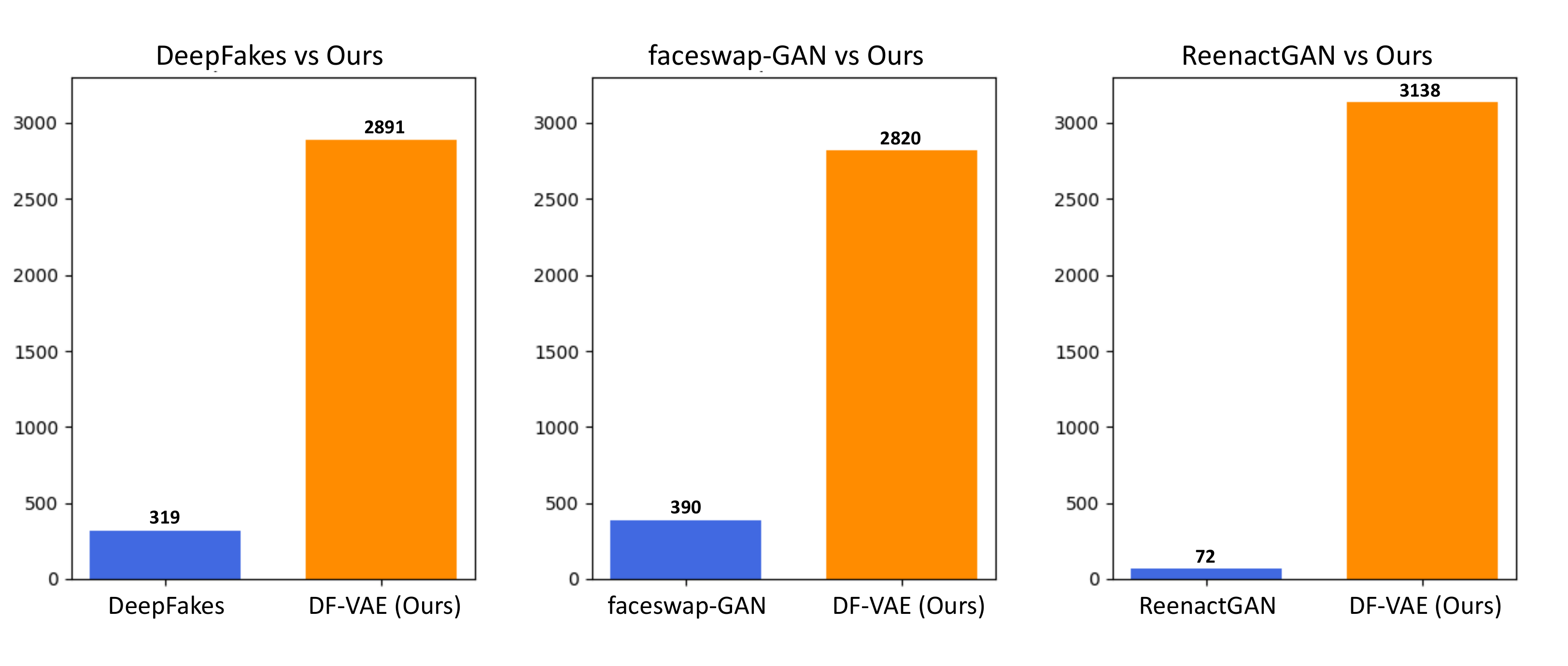}
   \end{center}
   \vspace{-0.35cm}
      \caption{Results of user study comparing methods. The bar charts show the number of users who give preference in each compared pair of manipulated videos.}
   \label{fig:user_study_method}
\end{figure}

\noindent
\textbf{Results.} We randomly choose $30$ real videos from DeeperForensics-$1.0$ as the source videos and $30$ real videos from FaceForensics++ \cite{FF++data} as the target videos. Thus, each method generates $30$ fake videos. Same as the user study based on datasets, we conduct the user study based on methods among $100$ professional participants who specialize in computer vision research. Because there are corresponding fake videos, we let the users directly choose their preferred fake videos between those generated by other methods and those generated by DF-VAE. Finally, we got $3210$ answers for each compared pair. The results are shown in Figure~\ref{fig:user_study_method}. We can see that DF-VAE shows an impressive advantage over the baselines, underscoring the \textit{high quality} of DF-VAE-generated fake videos.

\section{Quantitative Evaluation Metrics}

\noindent
\textbf{Frechet Inception Distance (FID)} \cite{fid} is a widely exploited metric for generative models. FID evaluates the \textit{similarity} of distribution between the generated images and the real images. FID correlates well with the visual quality of the generated samples. A lower value of FID means a better quality.

\noindent
\textbf{Inception Score (IS)} \cite{is} is an early and somewhat widely adopted objective evaluation metric for generated images. IS evaluates two aspects of generation quality: \textit{articulation} and \textit{diversity}. A higher value of IS means a better quality.

\begin{table}[t]
\centering
\begin{tabular}{c|c|c}
\Xhline{1.1pt}
Method							&	FID             &   IS      \\
\hline
DeepFakes \cite{DeepFakes}   	&	25.771          &   1.711   \\
faceswap-GAN \cite{faceswap-GAN} &	24.718          &   1.685   \\
ReenactGAN \cite{reenactgan}  	&	26.325          &   1.690   \\
\hline
\textbf{DF-VAE (Ours)}			&	\textbf{22.097} &   \textbf{1.714}   \\
\Xhline{1.1pt}
\end{tabular}
\caption{The FID and IS scores of DeepFakes \cite{DeepFakes}, faceswap-GAN \cite{faceswap-GAN}, ReenactGAN \cite{reenactgan}, and DF-VAE (Ours).}\label{tab:fidis}
\end{table}

Table~\ref{tab:fidis} shows the FID and IS scores of our method compared to other methods. DF-VAE outperforms all the three baselines in quantitative evaluations by FID and IS.

%


\section{Ablation Study}
\begin{figure*}
   \begin{center}
       \includegraphics[width=\linewidth]{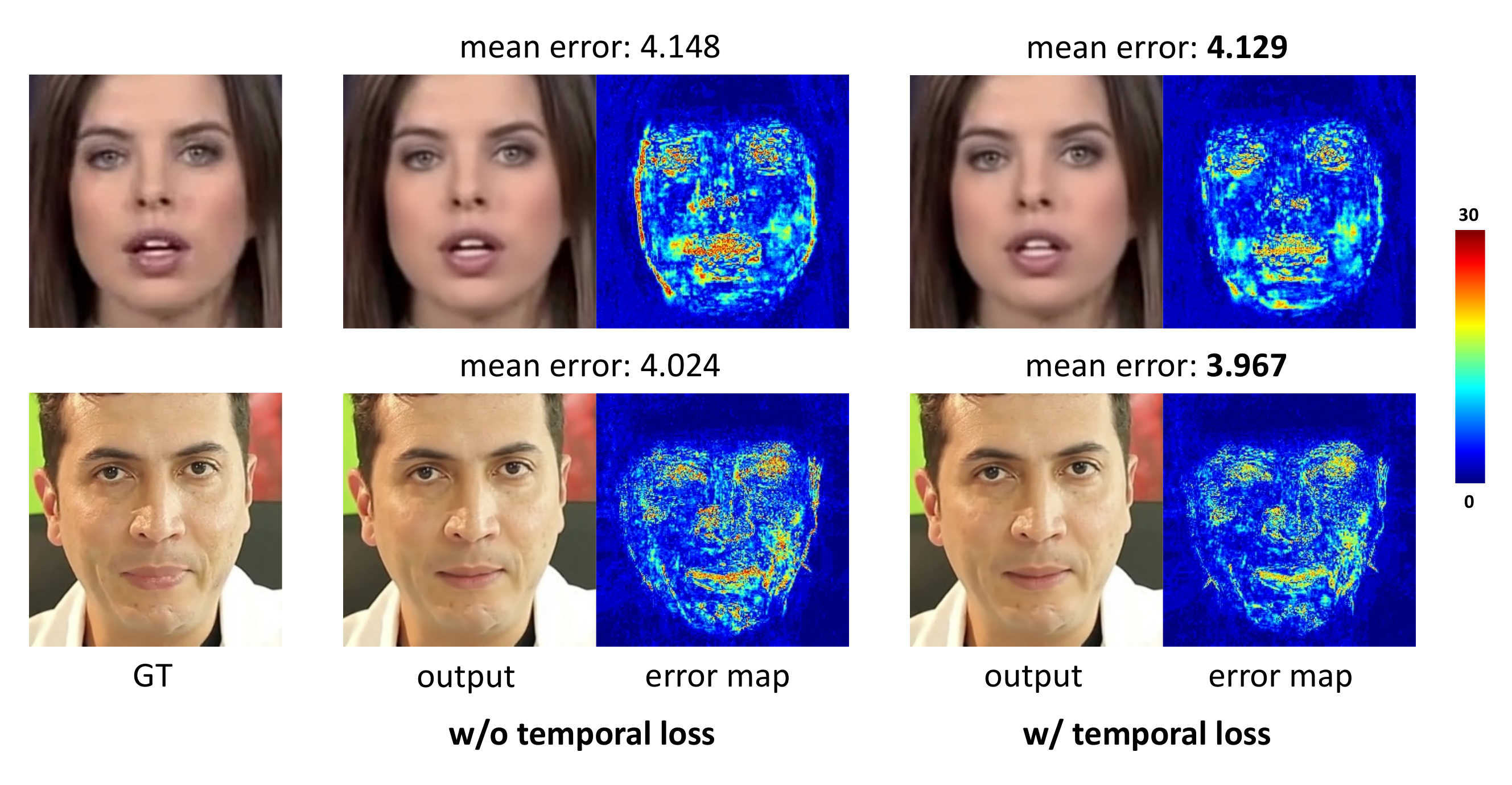}
   \end{center}
   \vspace{-0.15cm}
      \caption{The quantitative evaluation of the effectiveness of temporal loss. Similar to \cite{deepvideo}, we use the re-rendering error in a self-reenactment setting, where the ground truth is known. The error maps show Euclidean distance of per pixel in RGB channels ([$0$, $255$]). The mean errors are shown above the images. The corresponding color scale \textit{w.r.t.} error values is shown on the right side of the images.}
   \label{fig:temporal}
\end{figure*}

\noindent
\textbf{Ablation study of temporal loss.}
Since the swapped faces do not have the ground truth, we evaluate the effectiveness of temporal consistency constraint, \ie, temporal loss, in a self-reenactment setting. Similar to \cite{deepvideo}, we quantify the re-rendering error by Euclidean distance of per pixel in RGB channels ([$0$, $255$]). Visualized results are shown in Figure~\ref{fig:temporal}. Without the temporal loss, the re-rendering error is higher, hence demonstrating the effectiveness of temporal consistency constraint.

\begin{figure*}
   \begin{center}
       \includegraphics[width=\linewidth]{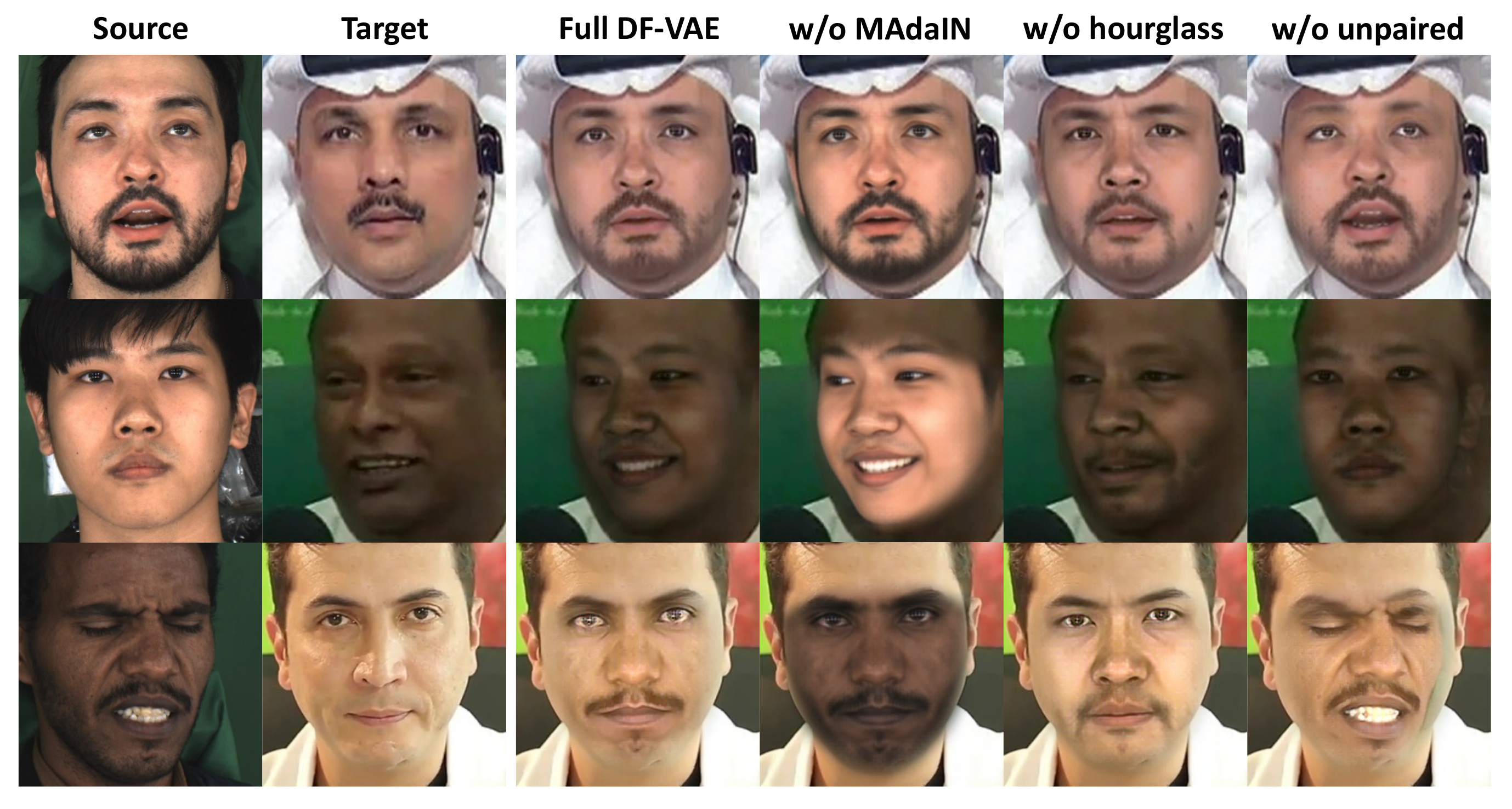}
   \end{center}
      \caption{The ablation studies of different components in DF-VAE framework in the many-to-many face swapping setting. Column $1$ and Column $2$ show the source face and the target face, respectively. Column $3$ shows the results of the full method. Column $4$, $5$, $6$ show the results when removing MAdaIN module, hourglass (structure extraction) module, and unpaired data construction, respectively.}
   \label{fig:ablation}
\end{figure*}

\noindent
\textbf{Ablation study of different components.}
We conduct further ablation studies \textit{w.r.t.} different components of our DF-VAE framework under many-to-many face swapping setting (see Figure~\ref{fig:ablation}). The source and target faces are shown in Column $1$ and Column $2$. In Column $3$, our full method, DF-VAE, shows high-fidelity face swapping results. In Column $4$, style mismatch problems are very obvious if we remove the MAdaIN module. If we remove the hourglass (structure extraction) module, the disentanglement of structure and appearance is not very thorough. The swapped face will be a mixture of multiple identities, as shown in Column $5$. When we perform face swapping without constructing unpaired data in the same domain (see Column $6$), the disentangled module will completely reconstruct the faces on the side of $E_\beta$, thus the disentanglement is not established at all. Therefore, the quality of face swapping will degrade if we remove any component in DF-VAE framework.

\section{Details of Benchmark Baselines}
\label{baselines}
We will elaborate on five baselines used in our face forgery detection benchmark in this section. Our benchmark contains four video-level face forgery detection methods, C3D \cite{c3d}, Temporal Segment Networks (TSN) \cite{tsn}, Inflated 3D ConvNet (I3D) \cite{i3d}, and ResNet+LSTM \cite{resnet, lstm}. One image-level detection method, XceptionNet \cite{xception}, which achieves the best performance in FaceForensics++ \cite{FF++data}, is evaluated as well.

\begin{itemize}
    \item \textbf{C3D} \cite{c3d} is a simple but effective method, which incorporates 3D convolution to capture the spatiotemporal feature of videos. It includes $8$ convolutional, $5$ max-pooling, and $2$ fully connected layers. The size of the 3D convolutional kernels is $3 \times 3 \times 3$. When training C3D, the videos are divided into non-overlapped clips with 16-frames length, and the original face images are resized to $112 \times 112$.
    \item \textbf{TSN} \cite{tsn} is a 2D convolutional network, which splits the video into short segments and randomly selects a snippet from each segment as the input. The long-range temporal structure modeling is achieved by the fusion of the class scores corresponding to these snippets. In our experiment, we choose BN-Inception \cite{BN} as the backbone and only train our model with the RGB stream. The number of segments is set to $3$ as default, and the original images are resized to $224 \times 224$.

    \item \textbf{I3D} \cite{i3d} is derived from Inception-V1 \cite{BN}. It inflates the 2D ConvNet by endowing the filters and pooling kernels with an additional temporal dimension. In the training, we use $64$-frame snippets as the input, whose starting frames are randomly selected from the videos. The face images are resized to $224 \times 224$.

    \item \textbf{ResNet+LSTM} \cite{resnet, lstm} is based on ResNet \cite{resnet} architecture. As a 2D convolutional framework, ResNet \cite{resnet} is used to extract spatial features (the output of the last convolutional layer) for each face image. In order to encode the temporal dependency between images, we place an LSTM \cite{lstm} module with $512$ hidden units after ResNet-$50$ \cite{resnet} to aggregate the spatial features. An additional fully connected layer serves as the classifier. All the videos are downsampled with a ratio of $5$, and the images are resized to $224 \times 224$ before feeding into the network. During training, the loss is the summation of the binary entropy on the output at all time steps, while only the output of the last frame is used for the final classification in inference.

    \item \textbf{XceptionNet} \cite{xception} is a depthwise-separable-convolution based CNN, which has been used in \cite{FF++data} for image-level face forgery detection. We exploit the same XceptionNet model as \cite{FF++data} but without freezing the weights of any layer during training. The face images are resized to $299 \times 299$. In the test phase, the prediction is made by averaging classification scores of all frames within a video.
\end{itemize}

\section{More Examples of Data Collection}
\label{datacollectionmore}
\begin{figure*}[t]
   \begin{center}
       \includegraphics[width=\linewidth]{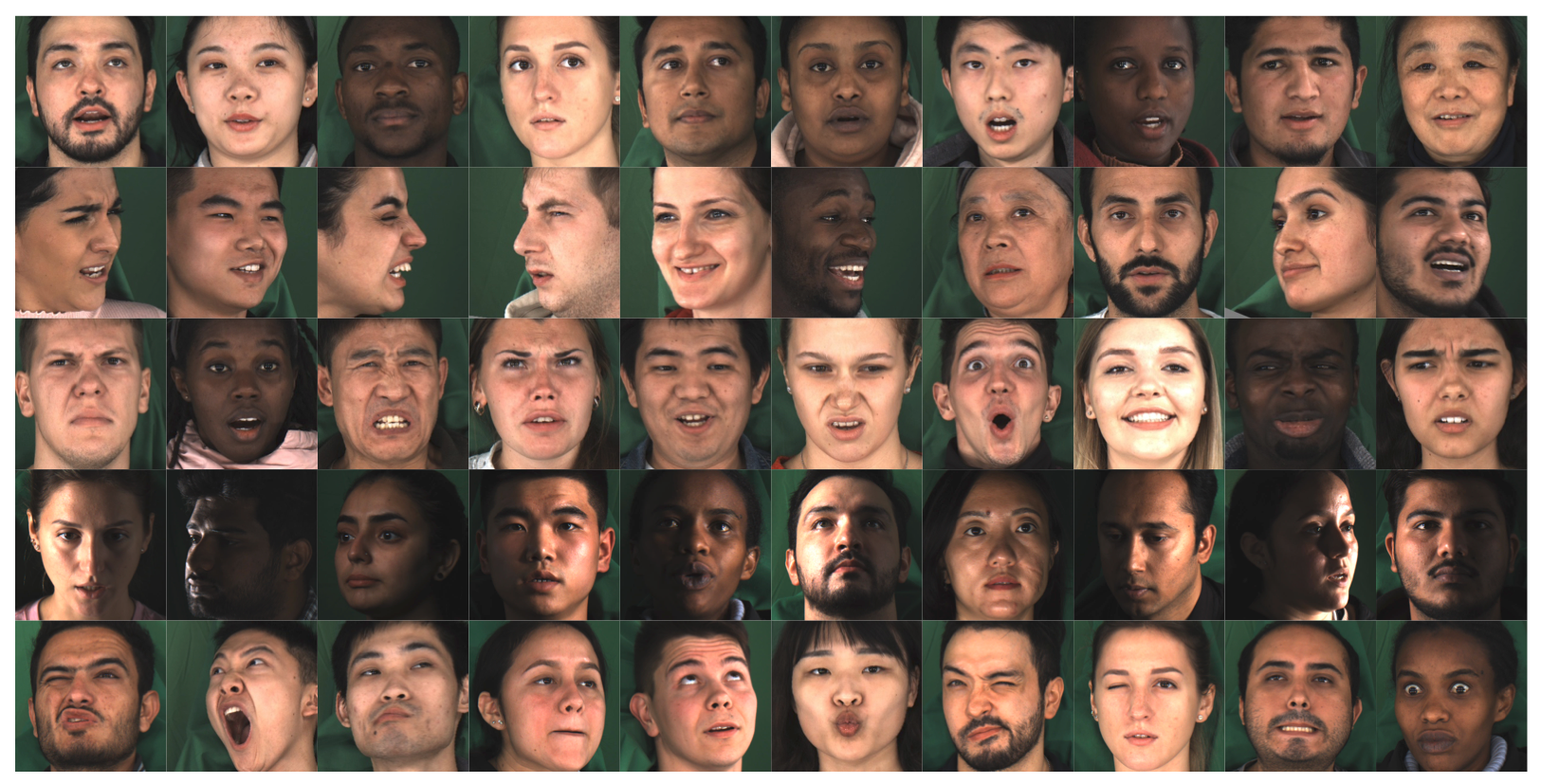}
   \end{center}
      \caption{More examples of the source video data collection. Our high-quality collected data vary in identities, poses, expressions, emotions, lighting conditions, and 3DMM blendshapes \cite{3dmm}.}
   \label{fig:supp_data_collection}
   \vspace{-0.75cm}
\end{figure*}

In this section, we will show more examples of our extensive source video data collection (see Figure~\ref{fig:supp_data_collection}). Our high-quality collected data vary in identities, poses, expressions, emotions, lighting conditions, and 3DMM blendshapes \cite{3dmm}. The source videos will also be released for further research.

\section{Perturbations}
\label{perturbations}
\begin{figure*}
   \begin{center}
       \includegraphics[width=\linewidth]{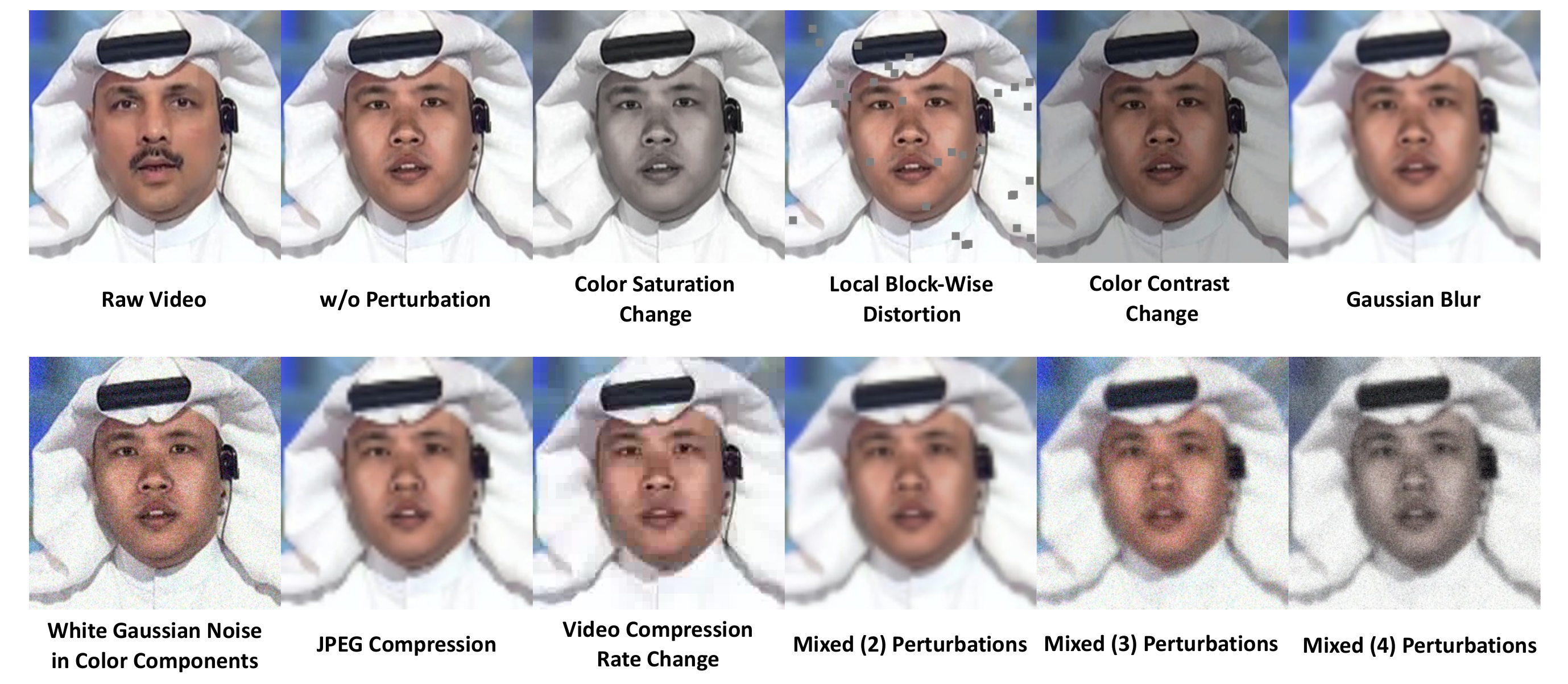}
   \end{center}
      \caption{Seven types of perturbations and the mixture of two (Gaussian blur, JPEG compression) / three (Gaussian blur, JPEG compression, white Gaussian noise in color components) / four (Gaussian blur, JPEG compression, white Gaussian noise in color components, color saturation change) perturbations in DeeperForensics-$1.0$.}
   \label{fig:perturbations}
\end{figure*}

We will also show some examples of perturbations in DeeperForensics-$1.0$. Seven types of perturbations and the mixture of two (Gaussian blur, JPEG compression) / three (Gaussian blur, JPEG compression, white Gaussian noise in color components) / four (Gaussian blur, JPEG compression, white Gaussian noise in color components, color saturation change) perturbations are shown in Figure~\ref{fig:perturbations}. These perturbations are very common distortions existing in real life.  The comprehensiveness of perturbations in DeeperForensics-$1.0$ ensures its \textit{diversity} to better simulate fake videos in real-world scenarios.

\end{document}